%% file: main.tex
\documentclass[11pt]{article}

\usepackage[final]{acl}

\usepackage{latexsym}
\usepackage{microtype}

\usepackage{fontspec}
\usepackage[bidi=default]{babel}
\babelprovide[main, import]{english}
\babelprovide[import]{arabic}
\babelfont[arabic]{rm}[
  Extension = .ttf,
  UprightFont = Amiri-Regular,
  BoldFont = Amiri-Bold,
  ItalicFont = Amiri-Italic,
  BoldItalicFont = Amiri-BoldItalic,
  Script=Arabic
]{Amiri}

\usepackage{inconsolata}

\usepackage{float}
\usepackage{todonotes}
\usepackage{xcolor}
\usepackage[switch,mathlines]{lineno}
\usepackage{etoolbox}
\usepackage{url}

\usepackage{graphicx}
\usepackage{booktabs}
\usepackage{multirow}
\usepackage{colortbl}
\usepackage{xcolor}
\usepackage{lipsum}
\usepackage[shortlabels]{enumitem}
\usepackage{subcaption}
\usepackage{tcolorbox}
\usepackage{fontawesome5}

\newenvironment{arab}%
  {\null\hfill\begingroup\selectlanguage{arabic}}%
  {\endgroup}
\definecolor{headerpink}{HTML}{FADBD8}
\definecolor{headergold}{HTML}{FCF3CF}
\definecolor{headergreen}{HTML}{D6FFC1}
\definecolor{headerblue}{HTML}{D6EAF8}

\title{Cultural Benchmarking of LLMs in Standard and Dialectal Arabic Dialogues}

\author{Muhammad Dehan Al Kautsar\thanks{Equal contribution.}\textsuperscript{1} \quad 
Saeed Almheiri$^*$\textsuperscript{1} \quad
Momina Ahsan$^*$\textsuperscript{1} \\ 
\textbf{Bilal Elbouardi$^*$\textsuperscript{1}}\quad 
\textbf{Younes Samih\textsuperscript{2}}\quad
\textbf{Sarfraz Ahmad\textsuperscript{1}}\quad 
\textbf{Amr Keleg\textsuperscript{1}} \\ 
\textbf{Omar El Herraoui\textsuperscript{1}}\quad 
\textbf{Kareem Elzeky\textsuperscript{1}}\quad 
\textbf{Abed Alhakim Freihat\textsuperscript{1}}\quad
\textbf{Mohamed Anwar\textsuperscript{1}}\\
\textbf{Zhuohan Xie\textsuperscript{1}}\quad 
\textbf{Junhong Liang\textsuperscript{1}}\quad 
\textbf{Mohammad Rustom Al Nasar\textsuperscript{3}}\\
\textbf{Preslav Nakov\textsuperscript{1}}\quad 
\textbf{Fajri Koto\textsuperscript{1}} \\
\textsuperscript{1}Mohamed bin Zayed University of Artificial Intelligence \\
\textsuperscript{2}IBM Research AI\quad
\textsuperscript{3}American University in the Emirates \\
\parbox{\linewidth}{\centering
\texttt{\{muhammad.dehan, saeed.y, momina.ahsan, bilal.elbouardi\}@mbzuai.ac.ae} \\
\faDatabase\ \href{https://huggingface.co/datasets/Almheiri/ArabCulture-Dialogue}{ArabCulture-Dialogue}}
} 

\begin{document}
\maketitle

\input{sections/0_abstract}
\input{sections/1_introduction}
\input{sections/2_related_work}
\input{sections/3_data}
\input{sections/4_experiment}
\input{sections/5_conclusion}
\input{sections/6_limitations}
\bibliography{ref}
\pagebreak
\input{sections/7_appendix}

\end{document}

%% file: sections/0_abstract.tex
\begin{abstract}
    
There is a significant gap in evaluating cultural reasoning in LLMs using conversational datasets that capture culturally rich and dialectal contexts. Most Arabic benchmarks focus on short text snippets in Modern Standard Arabic (MSA), overlooking the cultural nuances that naturally arise in dialogues. To address this gap, we introduce \mbox{\textbf{ArabCulture-Dialogue}}, a culturally grounded conversational dataset covering 13 Arabic-speaking countries, in both MSA and each country's respective dialect, spanning 12 daily-life topics and 54 fine-grained subtopics. We utilize the dataset to form three benchmarking tasks: (i) multiple-choice cultural reasoning, (ii) machine translation between MSA and dialects, and (iii) dialect-steering generation. Our experiments indicate that the performance gap between MSA and Arabic dialects still exists, whereby the models perform worse on all three tasks in the dialectal setup, compared to the MSA one.

\end{abstract}

%% file: sections/1_introduction.tex
\section{Introduction}

\begin{figure}[!t]
    \centering
    \includegraphics[width=0.85\linewidth]{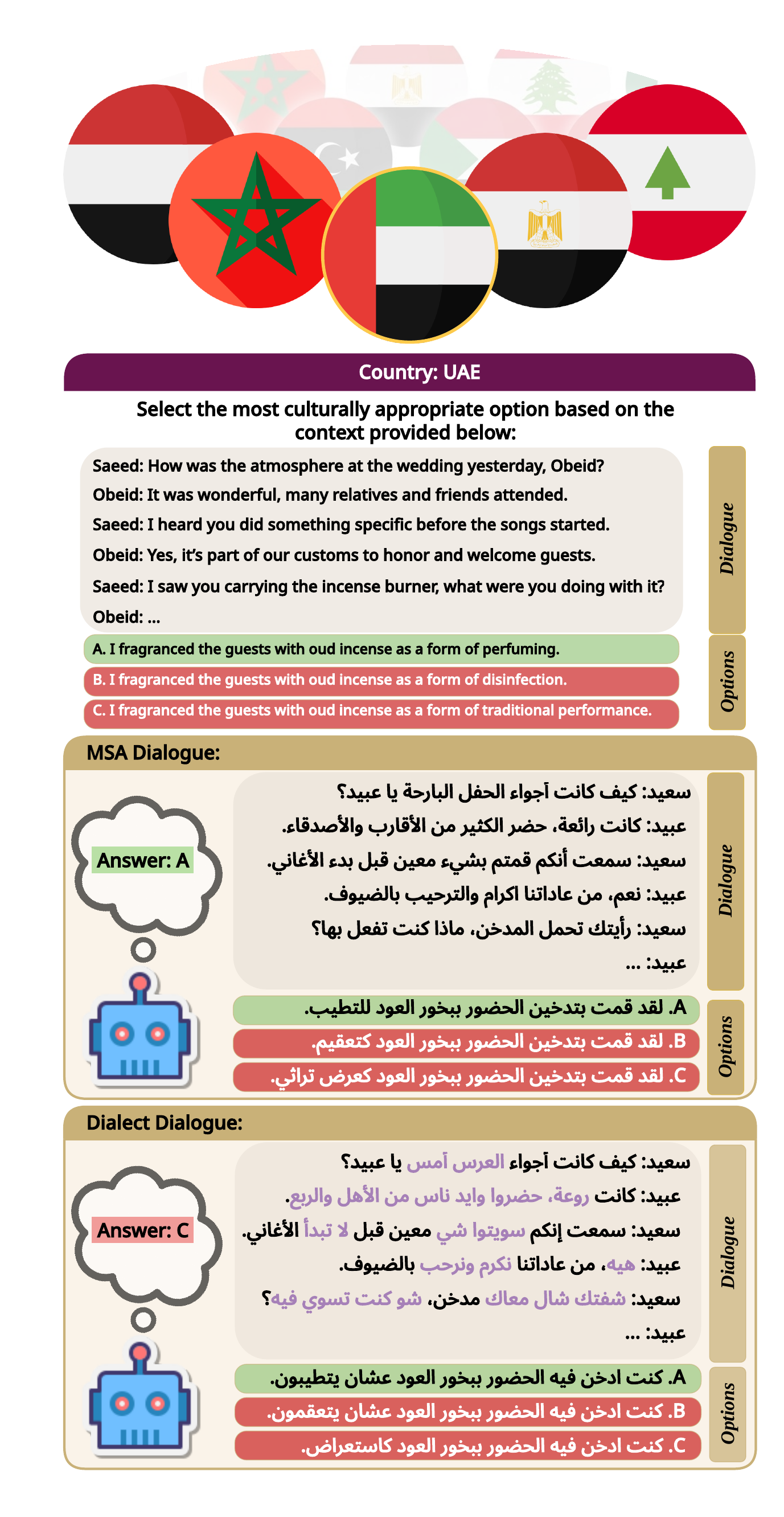} 
    \caption{Example dialogue from \textbf{ArabCulture-Dialogue}, related to weddings in the UAE, in both MSA and Emirati (UAE's) dialect. The English translation is provided for clarity, and is not part of the dataset.}
    \label{fig:intro}
\end{figure}

Arabic has over 400 million speakers, making it one of the most widely used languages in the world \citep{unesco_world_arabic_language_day}. While Modern Standard Arabic (MSA) serves as the formal written standard, most everyday communication occurs in diverse regional dialects that vary widely across and within countries \citep{habash2010introduction}. These dialects differ from MSA phonologically, lexically, grammatically, and pragmatically, encoding culturally grounded norms and practices. They are also shaped by local histories, language contact, and migration, contributing to variation even within the same country. For most speakers, dialect is the primary medium for expressing and transmitting cultural knowledge in daily conversation \citep{KWAIK20182}.

Recent years have seen substantial progress in Arabic NLP, with the emergence of Arabic-centric LLMs such as Jais \citep{sengupta2023jais}, SILMA \citep{silma-9b-2024}, and ALLaM \citep{bari2025allam}, alongside multilingual models that increasingly support Arabic. Evaluation benchmarks also have expanded accordingly. 

Among these, cultural commonsense reasoning has emerged as a particularly important dimension, as it probes whether models can reason about the shared knowledge, customs, and social expectations that underlie human communication. ArabCulture \citep{sadallah-etal-2025-commonsense} is a notable example, providing a manually–created benchmark of 3,482 questions across 13 countries and 54 cultural topics. However, existing cultural reasoning benchmarks, including ArabCulture, rely exclusively on isolated, single-turn multiple-choice questions presented in MSA. This evaluation paradigm, while useful for controlled assessment, diverges fundamentally from how cultural knowledge is actually exchanged and applied. In natural settings, cultural reasoning unfolds across conversational turns, where speakers must interpret implicit norms, respond appropriately to culturally situated utterances, and maintain pragmatic coherence throughout an interaction. Moreover, such exchanges are expected to be in dialects, suggesting that current benchmarks may systematically overestimate model capabilities by evaluating in a register that is both simpler and less culturally laden than authentic usage. This raises a critical question: \textit{can models that perform adequately on MSA-based cultural questions actually apply this knowledge in natural and dialect-mediated dialogue?}

To address this gap, we introduce \textbf{ArabCulture-Dialogue}, a human-curated conversational dataset that extends ArabCulture into multi-turn dialogues in both MSA and country-specific dialects. As illustrated in \autoref{fig:intro}, each instance consists of a culturally grounded conversation followed by three candidate responses, only one of which is culturally appropriate in both MSA and the local dialect. To our knowledge, this is the first dataset to benchmark Arabic cultural commonsense reasoning in a dialogue-based setting across MSA and regional dialects.

We also define three evaluation tasks on ArabCulture-Dialogue: (i) dialogue-based multiple choice cultural reasoning, which requires selecting the culturally appropriate response from three answer options; (ii) dialect translation between MSA and country-specific varieties; and (iii) dialect steering, which tests controlled generation in a specified dialect. Together, these tasks evaluate cultural reasoning in context, cross-register linguistic competence, and dialect-aware generation, while assessing how well models adapt to different settings and maintain consistency across tasks.

We evaluate a range of Arabic-centric, multilingual, and proprietary LLMs. Results show consistent degradation in performance on dialectal dialogues compared to MSA, with smaller open-weight models performing especially poorly. Cultural reasoning in MSA often fails to transfer to dialectal settings, and fine-grained country-level knowledge remains difficult. These findings highlight substantial limitations in current LLMs for culturally grounded, dialect-rich Arabic dialogue.

Our contributions are threefold: 
\begin{enumerate}[itemsep=1pt, topsep=3pt]
    \item We introduce \textbf{ArabCulture-Dialogue}, the first parallel MSA–dialect cultural dialogue dataset covering 13 Arab countries, created through rigorous human curation by 26 native speakers.
    \item We define three evaluation tasks: cultural MCQ, dialect translation, and dialect steering, to comprehensively assess culturally grounded dialogue capabilities.
    \item We conduct extensive experiments showing that dialectal cultural reasoning remains challenging for current open models, highlighting the need for culturally aware systems supporting dialectal inputs.
\end{enumerate} 

%% file: sections/2_related_work.tex
\section{Related Work}

\paragraph{Dialect and Cultural Reasoning in NLP:} Dialectal variation often encodes culturally grounded meaning beyond surface-level linguistic differences. Studies on English, Hindi, and Chinese dialects show that dialect choice signals social identity, politeness, norms, power relations, and pragmatic conventions \cite{hovy-2015-demographic, blodgett2016demographic, jurgens2017incorporating, hershcovich-etal-2022-challenges}. These effects are often context-dependent and become more apparent in interaction rather than isolated utterances. Despite this, many NLP approaches historically treat dialects as noise to be normalized toward a standard variety, causing large language models to degrade in performance and exhibit bias on dialectal inputs \cite{hofmann2024dialect, cao-etal-2023-assessing}. These findings highlight the need for culturally grounded evaluation. Yet, existing benchmarks rarely capture dialectal cultural reasoning in interactive settings. Our work addresses this gap by evaluating cultural reasoning in dialogue, where dialect-mediated norms emerge across turns rather than isolated prompts.

\paragraph{Arabic and Dialectal NLP:}
Arabic presents an informative case due to its diglossic nature: Modern Standard Arabic (MSA) dominates formal writing, education, and most NLP benchmarks, while everyday communication across the Arab world occurs primarily in regional dialects. These dialects encode region-specific idioms, politeness, humor, and social norms \cite{holes2006arabic}, often absent in MSA, making dialect choice closely tied to cultural identity and pragmatic intent \cite{abdulmageed2021nadi, bouamor-etal-2018-madar}.

Despite this centrality in daily communication, most Arabic NLP resources have prioritized MSA due to its standardized orthography and data availability, treating dialects mainly as a technical challenge through identification, normalization, or conversion to MSA \cite{abdulmageed2021nadi, abdelali-etal-2021-qadi, zaidan-callison-burch-2014-arabic}. Our work instead evaluates cultural reasoning without collapsing dialectal input into MSA, allowing assessment of models' ability to interpret culturally meaningful dialectal cues in context.

\paragraph{Task-Specific Cultural Evaluation in Arabic:} Recent Arabic-specific benchmarks expose the limitations of MSA-centric and single-turn evaluation.\linebreak \texttt{ArabCulture} \cite{sadallah-etal-2025-commonsense}, \texttt{AraDiCE} \cite{mousi-etal-2025-aradice}, and \texttt{PALM} \cite{alwajih-etal-2025-palm} introduce culturally grounded Arabic benchmarks with prompts in MSA and local dialects, revealing substantial regional performance disparities even for strong models. While these datasets highlight the importance of culturally grounded evaluation in Arabic, they focus on single-turn settings, whereas our work extends this to \textit{multi-turn conversational interactions} requiring sustained cultural reasoning.

\paragraph{Conversational and Multimodal Cultural Resources:} Recent studies show that Arabic cultural reasoning becomes more challenging under realistic evaluation conditions. The PALM-X shared task \cite{alwajih-etal-2025-palmx} shows limited gains from task-specific fine-tuning, Beyond MCQ \cite{bhatti2025beyond} reports performance drops in open-ended and dialectal settings, and SaudiCulture \cite{ayash2025saudiculture} highlights challenges with fine-grained regional customs within a single country. The findings show that dialectal variation, open-ended generation, and cultural specificity expose limitations visible in simplified evaluations, motivating conversational and multimodal resources for cultural reasoning.

JAWAHER \cite{magdy2025jawaher} focuses on culturally grounded proverbs, NileCHAT \cite{elmikki2025nilechat} provides dialect-heavy conversational data, and benchmarks such as cuDialog \cite{cao-etal-2024-bridging}, Peacock \cite{alwajih2024peacock}, and JEEM \cite{kadaoui-etal-2026-jeem} show that cultural understanding often requires grounding across linguistic and visual modalities. While these efforts broaden cultural evaluation, they do not model how cultural norms are negotiated across conversational turns. Our work addresses this gap through multi-turn Arabic dialogue, where such norms emerge dynamically in context.

In summary, while prior work shows that dialects are central to Arabic cultural expression and that models struggle with dialectal inputs, existing benchmarks remain fragmented and largely single-turn; we address this gap with a multi-country conversational benchmark for evaluating cultural competence in realistic, multi-turn discourse.

%% file: sections/3_data.tex
\section{Dataset Construction}
\label{sec:dataset-construction}
 
\begin{figure*}[ht]
  \centering
  \includegraphics[width=2\columnwidth]{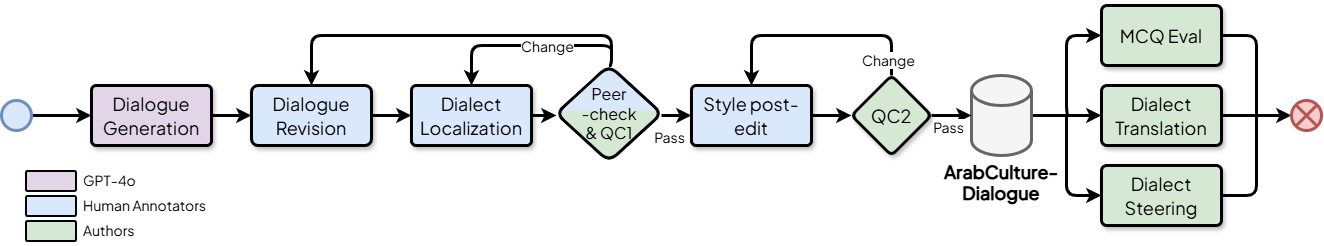}
  \caption{Dataset construction pipeline of \textbf{ArabCulture-Dialogue}. After the initial dialogue generation by GPT-5, all subsequent stages, including revision, dialect localization, style post-editing, and quality control, are performed through human annotation, resulting in a fully human-curated dataset.}
  \label{fig:dataset-construction-pipeline}
\end{figure*}

We construct a human-curated, culturally grounded dialogue dataset by transforming the ArabCulture benchmark \cite{sadallah-etal-2025-commonsense} into multi-turn conversations in both MSA and 13 Arabic dialects using the pipeline in \autoref{fig:dataset-construction-pipeline}. ArabCulture provides culturally relevant scenarios with one correct and two incorrect continuations. We preserve its country distribution and subtopic coverage, using each instance as a basis for creating richer conversational data.

\subsection{From Cultural Premises to MSA Dialogues}

For each ArabCulture sample, we first generate a short MSA dialogue based on the original premise and answer descriptions, with three potential continuations, only one of which is culturally sound. GPT-4o produces an initial draft, which is then manually revised by two native Arabic speakers from the corresponding country.\footnote{All annotators were compensated fairly, and the dataset creation cost was approximately USD 10K.} Samples are split equally between annotators within each country, all of whom are required to be native, familiar with local cultural norms, and fluent in both MSA and the local dialect. The use of large language models by annotators is strictly prohibited throughout the data construction pipeline. Details of annotator requirements are provided in Appendix~\ref{sec:appendix-annotator}.

During revision, the annotators verify the linguistic correctness, naturalness, and cultural appropriateness of the dialgoues. They also ensure internal consistency while addressing two common issues identified during early inspection: 
(1) \textbf{information leakage}, where the dialogue explicitly reveals the correct answer, and 
(2) \textbf{stylistic cues}, where the correct answer noticeably differs in tone or structure from the incorrect ones. 
Hence, selecting the correct option requires genuine cultural reasoning rather than reliance on superficial patterns.

\subsection{Dialect Localization and First Quality Check}

Each revised MSA dialogue is translated into the dialect of the corresponding country by the annotators who revised the dialogue. Annotators are instructed to avoid literal translation and instead produce natural, utterance-level conversational speech. Once dialect translation is completed by the two annotators, a different annotator performs an independent cross-review quality check (QC) of the translated dialogue, checking for dialect consistency, cultural correctness (including eliminating offensive content, if any), and adherence to the original MSA version. This multi-annotator workflow: MSA revision, dialect translation, and dialect cross-review, follows the formal guideline that is created for the annotators to help maintain consistent quality across all countries.

After the cross-annotator quality check is completed, we conduct an additional individual QC step. In this step, we randomly sample 50 instances per country and assess whether the dialogues meet the predefined quality criteria described above. If any instance fails to meet these standards, annotators are instructed to revise the dialogue or answer options accordingly. This QC process is carried out independently for each country, allowing the reviews to proceed in parallel and thereby improving efficiency.

\subsection{Post-Editing for Style Consistency and Second Quality Check}

During the first quality checks, we observe that some answer options were not stylistically aligned. For instance, the correct option might begin with a common discourse marker or be noticeably longer than the incorrect ones. These stylistic discrepancies can introduce unintended cues that make the correct answer easier to identify. Consequently, we introduce a post-editing stage, where annotators adjust all three answer options, in both MSA and dialect, to achieve comparable length, tone, and stylistic structure, while ensuring that only one option remains culturally correct. This step reduces unintentional stylistic cues and ensures that successful prediction relies on cultural reasoning rather than surface-level patterns.

After the answer-option refinement stage, we conduct a second round of quality check independently, without involving the original annotators, to ensure that the final dataset aligns with our intended goals. In this QC stage, we manually inspect 60 dialogues (30 in MSA and 30 in each dialect) per country and verify the following criteria: (1) Minimal stylistic differences exist among the three answer options, (2) The key information conveyed in the correct answer (as preserved from the original ArabCulture data) is retained, (3) Each answer option constitutes a natural and contextually appropriate response to the preceding dialogue (e.g., responses appropriately address preceding questions), and (4) The edited MSA answer options and their dialect counterparts are parallel in content and intent. Based on our evaluation, almost all samples from each country pass these criteria, with only one or two samples exhibiting minor, non-critical issues. Finally, we merge and finalize the validated instances to construct the first parallel MSA-dialect cultural dialogue dataset, which we refer to as \textbf{ArabCulture-Dialogue}. 

The entire data construction pipeline involves contributions from 26 annotators in total, resulting in a fully human-curated dataset designed to preserve both conversational quality and cultural authenticity. This collaborative process ensures coverage across countries and dialects while maintaining consistency in annotation standards.

\subsection{\textbf{ArabCulture-Dialogue}}

Through a carefully designed data construction pipeline, i.e., comprising generation, human revision, dialect translation, post-editing, and several quality checks, we produce a parallel MSA–dialect dialogue dataset that preserves the cultural grounding of ArabCulture while introducing a richer conversational context. This dataset provides a strong foundation for evaluating cultural reasoning, translation, and dialect-aware language generation in large language models. To the best of our knowledge, \textbf{ArabCulture-Dialogue} is the first dataset to benchmark Arabic cultural commonsense grounding across both MSA and 13 Arabic dialects within a dialogue-based setting, where cultural interactions are naturally expressed.

Each final instance in our dataset consists of an MSA dialogue, its localized dialect version, and three answer options written in both MSA and the corresponding dialect. \autoref{tab:data-stats} presents an overview of the \textbf{ArabCulture-Dialogue} dataset, while Appendix~\ref{sec:appendix-data-stats-region-topic} provides detailed statistics broken down by country and topic. Overall, the MSA portion of \textbf{ArabCulture-Dialogue} contains more total tokens than the dialect portion, whereas the dialectal data has a larger vocabulary size. The latter could be attributed to the fact that Arabic dialects have no standardized orthography.

\setlength{\tabcolsep}{0.2em}
\begin{table}[t]
\centering
\small
\begin{tabular}{l | c c c}
\toprule
\textbf{Statistic} & \textbf{MSA} & \textbf{Dialect} & \textbf{Merged} \\
\midrule
\#dialogues & 3,471 & 3,471 & 6,942 \\
\#country-specific dialogue & 1,390 & 1,390 & 2,780 \\
avg. words per dialogue & 50.57 & 48.48 & 49.53 \\
avg. utterance per dialogue & 6.06 & 6.06 & 6.06 \\
avg. words per utterance & 7.36 & 7.01 & 7.18 \\
\#words & 175,515 & 168,289 & 343,804 \\
\#unique words & 22,116 & 30,894 & 41,109 \\
\midrule 
\#unique words & \multicolumn{3}{c}{11,901 (38.52\% of} \\
(MSA $\wedge$ Dialect) & \multicolumn{3}{c}{Dialect unique words)} \\
\bottomrule
\end{tabular}
\caption{\textbf{ArabCulture-Dialogue}'s statistics. \textit{MSA} and \textit{Dialect} refer to the two subsets of the dataset, while \textit{Merged} denotes the full aggregated dataset.}
\label{tab:data-stats}
\end{table}

\setlength{\tabcolsep}{0.4em}
\setlength{\arrayrulewidth}{1.1pt}
\begin{table*}[bt]
\centering
\small
\begin{tabular}{l | ccc | ccc | ccc | ccc}
\toprule
\multirow{3}{*}{\textbf{Model}} & \multicolumn{6}{c|}{\textbf{Context: None}} & \multicolumn{6}{c}{\textbf{Context: Region + Country}}\\
\cline{2-13}
& \multicolumn{3}{c|}{\textbf{MSA}} & \multicolumn{3}{c|}{\textbf{Dialect}} & \multicolumn{3}{c|}{\textbf{MSA}} & \multicolumn{3}{c}{\textbf{Dialect}} \\
 & \textbf{Acc} & \textbf{CS} & \textbf{\textasciitilde CS} & \textbf{Acc} & \textbf{CS} & \textbf{\textasciitilde CS} & \textbf{Acc} & \textbf{CS} & \textbf{\textasciitilde CS} & \textbf{Acc} & \textbf{CS} & \textbf{\textasciitilde CS} \\
\midrule
Random & 0.333 & 0.333 & 0.333 & 0.333 & 0.333 & 0.333 & 0.333 & 0.333 & 0.333 & 0.333 & 0.333 & 0.333 \\
\midrule
\rowcolor{headerpink} \multicolumn{13}{c}{\textbf{Proprietary Models (0-shot)}} \\
Gemini-2.5-pro & 0.942 & \textbf{\underline{0.920}} & 0.954 & 0.939 & 0.917 & 0.956 & 0.950 & \textbf{\underline{0.933}} & 0.961 & 0.939 & \textbf{\underline{0.924}} & 0.950 \\
GPT-5 & \textbf{\underline{0.943}} & 0.908 & \textbf{\underline{0.966}} & \textbf{\underline{0.948}} & \textbf{\underline{0.924}} & \textbf{\underline{0.964}} & \textbf{\underline{0.953}} & 0.927 & \textbf{\underline{0.971}} & \textbf{\underline{0.948}} & 0.923 & \textbf{\underline{0.965}} \\
\midrule
\rowcolor{headergreen} \multicolumn{13}{c}{\textbf{Arabic Centric Models (0-shot)}} \\
Jais-7B-chat & 0.554 & 0.504 & 0.589 & 0.498 & 0.460 & 0.524 & 0.531 & 0.492 & 0.558 & 0.469 & 0.440 & 0.489 \\
ALLaM-7B-Instruct & 0.418 & 0.387 & 0.437 & 0.398 & 0.374 & 0.413 & 0.506 & 0.465 & 0.533 & 0.471 & 0.440 & 0.491 \\
Cohere-Arabic-7B & 0.355 & 0.366 & 0.348 & 0.342 & 0.351 & 0.337 & 0.375 & 0.385 & 0.368 & 0.357 & 0.361 & 0.355 \\
Jais-2 8B-Chat & 0.740 & 0.689 & 0.774 & 0.710 & 0.650 & 0.751 & 0.766 & 0.704 & 0.808 & 0.731 & 0.667 & 0.773 \\
Fanar-1-9B & 0.391 & 0.373 & 0.403 & 0.350 & 0.354 & 0.348 & 0.618 & 0.553 & 0.662 & 0.534 & 0.486 & 0.566 \\
SILMA-9B-Instruct & \underline{{0.783}} & \underline{{0.715}} & \underline{{0.829}} & 0.716 & \underline{{0.663}} & 0.751 & 0.784 & 0.724 & 0.825 & 0.714 & 0.665 & 0.747 \\
Hala-9B & 0.779 & 0.711 & 0.826 & \underline{{0.733}} & 0.660 & \underline{{0.782}} & \underline{{0.820}} & \underline{{0.751}} & \underline{{0.866}} &  \underline{{0.763}} & \underline{{0.692}} & \underline{{0.810}} \\
\midrule
\rowcolor{headergold} \multicolumn{13}{c}{\textbf{Multilingual Models (0-shot)}} \\
Llama-3.1-8B-it & 0.456 & 0.424 & 0.478 & 0.412 & 0.406 & 0.416 & 0.473 & 0.441 & 0.495 & 0.430 & 0.422 & 0.435 \\
Qwen-3-8B & 0.354 & 0.350 & 0.357 & 0.354 & 0.346 & 0.359 & 0.380 & 0.368 & 0.387 & 0.368 & 0.355 & 0.377 \\
Gemma-2-9B-it & \underline{0.671} & \underline{0.619} & \underline{0.707} & \underline{0.609} & \underline{0.558} & \underline{0.644} & \underline{0.710} & \underline{0.661} & \underline{0.744} & \underline{0.643} & \underline{0.582} & \underline{0.684}  \\
\bottomrule
\end{tabular}
\caption{\textbf{Task 1 - MCQ Evaluation}: accuracies under two settings: \textit{None} (no context) and \textit{Region + Country}. Results are averaged and reported separately for country-specific (CS) and non-country-specific ($\sim$CS) dialogues. The best overall accuracy score is shown in \textbf{bold}, and the best score within each model category is \underline{underlined}.}
\label{tab:mcq-main-result}
\end{table*}

%% file: sections/4_experiment.tex
\section{Experimental Setup}

Using \textbf{ArabCulture-Dialogue}, we evaluate dialogue-based cultural commonsense reasoning in Arabic across (1) Arabic-centric large language models, (2) multilingual large language models, and (3) proprietary large language models. All model inferences are conducted using a single run.

The Arabic-centric models include Jais-Adapted-7B-Chat~\citep{sengupta2023jais}, Jais-2-8B-Chat~\citep{jaisfamilymodelcard}, ALLaM-7B-Instruct~\citep{bari2025allam}, SILMA-9B-Instruct~\citep{silma-9b-2024}, Cohere-Arabic-7B~\citep{alnumay2025command}, Fanar-1-9B~\citep{fanarllm2025}, and Hala-9B~\citep{hammoud2025halatechnicalreportbuilding}. The multilingual category includes Gemma-2-9B-Instruct~\citep{gemma_2024}, Qwen3-8B~\citep{qwen3technicalreport}, and LLaMA-3.1-8B-Instruct~\citep{grattafiori2024llama3herdmodels}. For proprietary models, we evaluate GPT-5 (with the reasoning level set to normal) and Gemini-2.5-Pro. To ensure a fair comparison, all Arabic-centric and multilingual models are constrained to a similar parameter scale, ranging from 7 billion to 9 billion parameters. In contrast, proprietary models are included to reflect the current state of the art.

Building on the manually curated parallel dialogue dataset described in the previous section, we evaluate these models across three complementary tasks: (1) dialogue-based cultural commonsense reasoning in multiple-choice question (MCQ) evaluation, (2) dialect translation, and (3) dialect steering. All prompts are written in English and provided in the Appendix~\ref{sec:appendix-task-prompts}.

\paragraph{Task 1 - MCQ Evaluation:}
For the MCQ evaluation task, models are presented with a dialogue and three answer options, only one of which is correct. We use the dataset in its original format, as it is already structured for this evaluation. In addition to this standard setting, we assess evaluation robustness by optionally providing explicit geographic context to the prompt (Region, or both Region and Country). Since cultural knowledge encoded in LLMs can vary across locations, this additional information might help the models better reason about culturally grounded dialogues~\citep{koto-etal-2024-indoculture,sadallah-etal-2025-commonsense}. The LLMs are evaluated using likelihood-based scoring rather than open-ended generation, bypassing stochastic sampling. We report accuracy scores that remain identical across multiple inference runs. Consequently, standard deviation is not reported.

\paragraph{Task 2 - Dialect Translation:} This task evaluates a model’s ability to translate multi-turn dialogues between Modern Standard Arabic (MSA) and country-specific Arabic dialects across 13 countries. Since the dataset contains parallel MSA–dialect dialogues for each country, the dialogues' corresponding utterances naturally form parallel translation pairs. For both translation directions (MSA to country-level dialect and country-level dialect to MSA), we specify the country's dialect and its respective region in the prompt.

We use the Arabic Level of Dialectness~(ALDi; \citealp{keleg-etal-2023-aldi}) to assess how dialectal an output is. ALDi is a continuous score that measures a sentence's divergence from MSA, with an ALDi score of zero implying that the sentence is in MSA. Moreover, we assess translation quality using BLEU~\citep{papineni-etal-2002-bleu_edit}, BERTScore with mBERT as the scoring model~\citep{devlin-etal-2019-bert}, and an LLM-as-a-judge framework based on GPT-5 with the reasoning level set to low (see the prompt in \autoref{fig:appendix-prompt-judge}). We evaluate all the models in the zero-shot setting and assess the impact of supervised fine-tuning on the performance of multilingual models.

\paragraph{Task 3 - Dialect Steering:} The task evaluates a model’s ability to control the dialectal variety of its generated responses. Given a dialogue context and an utterance in MSA, the model is instructed to produce a single response either in MSA or in a specified target dialect. 

This setting tests whether the model can both recognize and generate the intended dialect, which varies across countries. The model receives the dialogue context and completes it with one utterance in the target variety. We evaluate performance under both zero-shot and supervised fine-tuning settings, using an LLM-as-a-judge framework based on GPT-5. We also apply the GlotLID language identification model~\citep{kargaran-etal-2023-glotlid} to verify whether the generated output matches the target dialect automatically.

\section{Results and Analysis}
In this section, we report the results for the three tasks. Two observations apply to all of them:
\begin{enumerate}[leftmargin=*, noitemsep, nolistsep]
    \item Arabic-centric models outperform multilingual models of similar sizes.
    \item Large proprietary models are exceptional on the MCQ task, with gaps for the other tasks.
\end{enumerate}

\subsection{Task 1 - MCQ Evaluation}

Among the Arabic-centric models, Hala-9B and SILMA-9B perform better than the others, with Jais2-8B also remaining competitive, as shown in \autoref{tab:mcq-main-result}.
In contrast, the three smaller Arabic-centric models lag, even for relatively recent models such as ALLaM-7B and Cohere-Arabic-7B. This may suggest such models cannot perform dialogue-based cultural commonsense reasoning. However, other confounding factors exist.

\begin{table*}[hptb]
\centering
\small
\setlength{\tabcolsep}{4pt}
\begin{subtable}[]{0.99\textwidth}
\centering
\begin{tabular}{l | c | c | c | ccccc}
\toprule
\multirow{2}{*}{\textbf{Model}} &
\multicolumn{1}{c|}{\textbf{ALDi (0--1)}} &
\multicolumn{1}{c|}{\textbf{BLEU (0--1)}} &
\multicolumn{1}{c|}{\textbf{BERTScore}} &
\multicolumn{5}{c}{\textbf{LLM-as-Judge (1--5)}} \\
&
& 
B4 &
F1 &
Adeq. & Flu. & Reg. & Term. & Overall \\
\hline
Reference & 0.61\textsubscript{±0.3} & 1.000 & 1.000 & 4.131 & 4.003 & 4.256 & 4.320 & 3.862\\
\midrule
\rowcolor{headerpink}
\multicolumn{9}{c}{\textbf{Proprietary Models (0-shot)}} \\
Gemini-2.5-pro
& \textbf{0.67}\textsubscript{±0.3}
& 0.273\textsubscript{(2)}
& 0.877\textsubscript{(2)}
& 4.518 & 4.440 & 4.513 & 4.623 & 4.188\textsubscript{(2)} \\

GPT-5
& 0.61\textsubscript{±0.3}
& \textbf{0.276\textsubscript{(1)}}
& \textbf{0.879}\textsubscript{(1)}
& \textbf{4.915} & \textbf{4.704} & \textbf{4.651} & \textbf{4.925} & \textbf{4.530\textsubscript{(1)}} \\
\midrule

\rowcolor{headergreen}
\multicolumn{9}{c}{\textbf{Arabic-centric Models (0-shot)}} \\

Jais-7B-chat
& 0.11\textsubscript{±0.2}
& 0.129\textsubscript{(7)}
& 0.808\textsubscript{(7)}
& 4.021 & 3.528 & 1.482 & 4.184 & 2.439\textsubscript{(6)} \\

ALLaM-7B-Instruct
& \underline{0.46\textsubscript{±0.3}}
& \underline{0.196}\textsubscript{(3)}
& \underline{0.847}\textsubscript{(3)}
& \underline{4.199} & \underline{3.731} & \underline{3.019} & \underline{4.241} & \underline{3.408\textsubscript{(3)}} \\

Cohere-Arabic-7B
& 0.40\textsubscript{±0.3}
& 0.152\textsubscript{(5)}
& 0.835\textsubscript{(5)}
& 3.850 & 3.339 & 2.422 & 3.953 & 2.995\textsubscript{(4)} \\

Fanar-1-9B
& 0.44\textsubscript{±0.3}
& 0.156\textsubscript{(4)}
& 0.840\textsubscript{(4)}
& 4.007 & 3.214 & 2.206 & 3.986 & 2.943\textsubscript{(5)} \\

SILMA-9B-Instruct
& 0.39\textsubscript{±0.3}
& 0.035\textsubscript{(14)}
& 0.728\textsubscript{(14)}
& 1.443 & 1.492 & 1.278 & 1.576 & 1.302\textsubscript{(14)} \\

Hala-9B
& 0.06\textsubscript{±0.1}
& 0.122\textsubscript{(8)}
& 0.790\textsubscript{(11)}
& 2.870 & 2.829 & 1.060 & 3.403 & 1.764\textsubscript{(10)} \\

\midrule
\rowcolor{headergold}
\multicolumn{9}{c}{\textbf{Multilingual Models (0-shot)}} \\

Llama-3.1-8B-it
& \underline{0.51\textsubscript{±0.3}}
& 0.058\textsubscript{(12)}
& 0.787\textsubscript{(12)}
& 1.656 & 1.305 & 1.156 & 1.651 & 1.342\textsubscript{(13)} \\

Qwen-3-8B 
& 0.10\textsubscript{±0.2}
& \underline{0.115}\textsubscript{(10)}
& \underline{0.818}\textsubscript{(6)}
& \underline{3.418} & \underline{2.752} & \underline{1.512} & \underline{3.341} & \underline{2.354\textsubscript{(7)}} \\

Gemma-2-9B-it
& 0.53\textsubscript{±0.3}
& 0.071\textsubscript{(11)}
& 0.795\textsubscript{(9)}
& 1.884 & 1.408 & 1.306 & 1.708 & 1.495\textsubscript{(12)} \\
\midrule
\rowcolor{headerblue}
\multicolumn{9}{c}{\textbf{Multilingual Models (SFT)}} \\

Llama-3.1-8B
& 0.34\textsubscript{±0.3}
& 0.046\textsubscript{(13)}
& 0.747\textsubscript{(13)}
& 1.500 & 2.076 & 1.832 & 2.268 & 1.747\textsubscript{(11)} \\

Qwen-3-8B
& 0.14\textsubscript{±0.2}
& 0.122\textsubscript{(9)}
& 0.795\textsubscript{(10)}
& 2.054 & 2.269 & 1.562 & 2.998 & 2.040\textsubscript{(9)} \\

Gemma-2-9B-it
& \underline{0.41\textsubscript{±0.3}}
& \underline{0.135}\textsubscript{(6)}
& \underline{0.806}\textsubscript{(8)}
& \underline{2.071} & \underline{2.515} & \underline{2.185} & \underline{2.940} & \underline{2.210\textsubscript{(8)}} \\

\bottomrule
\end{tabular}

\caption{MSA to Dialect Translation.\\ \textbf{Note:} Dialectal Outputs are expected to have moderate to high ALDi scores.}

\label{tab:msa_dialect_main}
\end{subtable}

\hfill

\begin{subtable}[]{0.99\textwidth}
\centering
\begin{tabular}{l | c | c | c | ccccc}
\toprule
\multirow{2}{*}{\textbf{Model}} &
\multicolumn{1}{c|}{\textbf{ALDi (0--1)}} &
\multicolumn{1}{c|}{\textbf{BLEU (0--1)}} &
\multicolumn{1}{c|}{\textbf{BERTScore}} &
\multicolumn{5}{c}{\textbf{LLM-as-Judge (1--5)}} \\
& & B4 & F1 &
Adeq. & Flu. & Reg. & Term. & Overall \\
\midrule

Reference
& 0.03\textsubscript{±0.1} & 1.000 & 1.000 & 4.212 & 4.714 & 4.797 & 4.527 & 4.182 \\

\midrule
\rowcolor{headerpink}
\multicolumn{9}{c}{\textbf{Proprietary Models (0-shot)}} \\

Gemini-2.5-pro
& \textbf{0.03\textsubscript{±0.1}} & 0.430\textsubscript{(2)} & 0.909\textsubscript{(2)}
& 4.810 & 4.836 & 4.936 & 4.875 & 4.654\textsubscript{(2)} \\

GPT-5
& \textbf{0.03\textsubscript{±0.1}} & \textbf{0.434\textsubscript{(1)}} & \textbf{0.911\textsubscript{(1)}} & \textbf{4.905} & \textbf{4.883} & \textbf{4.957} & \textbf{4.927} & \textbf{4.773\textsubscript{(1)}} \\

\midrule
\rowcolor{headergreen}
\multicolumn{9}{c}{\textbf{Arabic-centric Models (0-shot)}} \\

Jais-7B-chat
& 0.51\textsubscript{±0.3} & 0.178\textsubscript{(13)} & 0.836\textsubscript{(13)}
& 4.447 & 2.422 & 1.563 & 3.870 & 2.299\textsubscript{(13)} \\

ALLaM-7B-Instruct
& 0.05\textsubscript{±0.1} & \underline{0.405\textsubscript{(3)}} & \underline{0.891\textsubscript{(3)}}
& \underline{4.253} & \underline{4.509} & \underline{4.666} & \underline{4.489} & \underline{4.079\textsubscript{(3)}} \\

Cohere-Arabic-7B
& 0.21\textsubscript{±0.3} & 0.310\textsubscript{(8)} & 0.873\textsubscript{(6)}
& 4.145 & 3.738 & 3.494 & 4.177 & 3.342\textsubscript{(6)} \\

Fanar-1-9B
& \textbf{0.03\textsubscript{±0.1}} & 0.367\textsubscript{(4)} & 0.881\textsubscript{(4)}
& 3.782 & 4.386 & 4.516 & 4.179 & 3.779\textsubscript{(4)} \\

SILMA-9B-Instruct
& 0.33\textsubscript{±0.3} & 0.218\textsubscript{(12)} & 0.841\textsubscript{(11)}
& 3.542 & 2.735 & 2.143 & 3.571 & 2.510\textsubscript{(10)} \\

Hala-9B
& 0.05\textsubscript{±0.1} & 0.315\textsubscript{(7)} & 0.839\textsubscript{(12)}
& 3.561 & 3.518 & 4.383 & 3.969 & 3.422\textsubscript{(5)} \\

\midrule
\rowcolor{headergold}
\multicolumn{9}{c}{\textbf{Multilingual Models (0-shot)}} \\

Llama-3.1-8B-it
& 0.13\textsubscript{±0.2} & 0.242\textsubscript{(11)} & 0.859\textsubscript{(8)}
& 2.660 & 2.461 & 2.576 & 2.805 & 2.468\textsubscript{(11)} \\

Qwen-3-8B
& 0.13\textsubscript{±0.2} & 0.265\textsubscript{(10)} & 0.868\textsubscript{(7)}
& 2.875 & 2.699 & 2.669 & 3.047 & 2.647\textsubscript{(9)} \\

Gemma-2-9B-it 
& \underline{0.08\textsubscript{±0.2}} & \underline{0.319\textsubscript{(5)}} & \underline{0.881\textsubscript{(5)}}
& \underline{3.423} & \underline{3.203} & \underline{3.417} & \underline{3.545} & \underline{3.128\textsubscript{(7)}} \\

\midrule
\rowcolor{headerblue}
\multicolumn{9}{c}{\textbf{Multilingual Models (SFT)}} \\

Llama-3.1-8B-it
& 0.05\textsubscript{±0.1} & 0.108\textsubscript{(14)} & 0.779\textsubscript{(14)}
& 1.666 & 2.483 & 3.088 & 2.502 & 2.040\textsubscript{(14)} \\

Qwen-3-8B
& 0.06\textsubscript{±0.1} & 0.285\textsubscript{(9)} & 0.845\textsubscript{(10)}
& 2.101 & 2.617 & 3.257 & 2.834 & 2.397\textsubscript{(12)} \\

Gemma-2-9B-it
& \underline{0.04\textsubscript{±0.1}} & \underline{0.316\textsubscript{(6)}} & \underline{0.847\textsubscript{(9)}}
& \underline{2.117} & \underline{3.117} & \underline{3.819} & \underline{3.151} & \underline{2.654\textsubscript{(8)}} \\

\bottomrule
\end{tabular}

\caption{Dialect to MSA Translation.\\ \textbf{Note:} MSA Outputs are expected to have zero ALDi scores.}
\label{tab:dialect_msa_main}

\end{subtable}

\caption{\textbf{Task 2 - Machine Translation}'s evaluation metrics with the models prompted with (Context: Country + Region). We report ALDi, BLEU, and BERTScore, along with LLM-as-Judge scores (1--5) for \textbf{Adeq.} (semantic adequacy), \textbf{Flu.} (fluency and grammaticality), \textbf{Reg.} (dialectal and regional correctness), \textbf{Term.} (terminology and lexical choice), and \textbf{Overall} (holistic quality). 
The best overall model is shown in \textbf{bold}, and the best model within each category is \underline{underlined}. The models' rankings according to each metric are reported as \textsubscript{(subscript)}. \textbf{Note:} We noticed a few outputs where the models generated extra outputs other than the dialogue translations.}
\end{table*}

As expected, most models have higher accuracy picking the right answer when fed with MSA dialogues than with their respective DA ones. However, the gap is not drastic.
For the dialogue's topics, all models almost categorically perform better on non-country-specific (\textasciitilde CS) dialogues than on country-specific (CS) ones. This is consistent with the results of the original ArabCulture benchmark from which our dialogue dataset was created \cite{sadallah-etal-2025-commonsense}. Lastly, providing information about the region and the country to which the dialogue is relevant in general increases the models' ability to pick the right answer.\footnote{Only providing the region as a context is also better than not, as shown in \autoref{tab:appendix-mcq-result} of Appendix~\ref{sec:appendix-mcq-eval-details}.}

\subsection{Task 2 - Dialect Translation}

As expected, the models perform better on translation to MSA (\autoref{tab:dialect_msa_main}) than to non-standardized dialects (\autoref{tab:msa_dialect_main}), according to the three evaluation metrics: BLEU, BERTScore, and LLM-as-judge ratings. Nevertheless, the performance gap could be exaggerated by the limitations of the different metrics, which are expected to work better for standardized and high-resource languages than non-standardized dialects. Further investigations are required to realistically estimate the gap.\footnote{We analyzed a sample of the LLM-as-judge ratings for Adequacy and Fluency in Appendix~\ref{sec:human_eval-agreement}, finding adequate agreement with manually-assigned ratings. However, we notice some non-negligible variation.}

Overall, it seems that the models' rankings according to each of the metrics are consistent. Moreover, the models' rankings across the two directions are similar, as elaborated below:

\paragraph{Proprietary Models Superiority:} GPT-5 and Gemini-2.5-Pro achieve the highest overall scores, according to BLEU, BERTScore, and the Overall score assigned by the LLM-as-a-Judge evaluation.

\paragraph{Capabilities of Open-weight Models:} 
Among Arabic-centric models, ALLaM-7B-Instruct is the strongest performer, followed by Fanar-1-9B, and Cohere-Arabic-7B. The corresponding ALDi scores and Register metric hint that the models are generally capable of generating valid translations (be it MSA or dialectal), but have some difficulty in using the correct dialectal forms and register.

Other models---such as Hala-9B and Qwen-3-8B---are capable of generating valid MSA translations, but fail to generate dialectal outputs as indicated by the low respective mean ALDi scores of 0.06 and 0.14, for MSA to dialect translation.

\paragraph{Impact of Fine-tuning on Multilingual Models:}

Supervised fine-tuning improves multilingual models in some dimensions but does not close the gap with zero-shot Arabic-centric models. The mean ALDi scores of the multilingual models' outputs indicate that they are capable of generating dialectal outputs. Manual inspection indicates that the translations are neither semantically correct nor using the desired dialect. Fine-tuned Gemma-2-9B-it shows moderate gains in BLEU and LLM-as-a-judge fluency and terminology, but register scores remain low. This suggests that limited supervised data is insufficient to robustly encode dialectal distinctions, especially across multiple countries.

\subsection{Task 3 - Dialect Steering}
\label{sec:dialect-steering}

We evaluate \emph{dialect steering} as a controlled generation task with two targets: Modern Standard Arabic (MSA) and country-dialect Arabic. For each prompt, we ask the model to continue a short dialogue either in MSA or in the target dialect, and score outputs with (i) an LLM-as-a-judge quality metric (1--5, reported as $(s{-}1)/4 \in [0,1]$) and (ii) dialect identity via GlotLID \citep{kargaran-etal-2023-glotlid}. GlotLID is reported in two ways: (1)~\emph{strict ISO-code accuracy} where exact ISO 639-3 match against the country target code, and (2)~\emph{macro-region accuracy} where a coarser mapping that collapses close dialects into Gulf/Levant/Nile River/North Africa, following \citet{bhatti2025beyond}.

\paragraph{Dialect steering overview:} \autoref{tab:dialect-steering-zero-shot-plus-sft} summarizes the performance of the different models. GPT-5 achieves the best judged quality for both MSA and dialect continuations, while Gemini-2.5-pro is slightly weaker on judged quality but noticeably stronger on strict-code GlotLID. Within Arabic-centric models, ALLaM-7B is the most reliable overall, whereas several Arabic-specialized baselines produce fluent continuations that nevertheless collapse toward wider regional varieties under strict ISO coding. The results hint that most models can respond in MSA. Moreover, some can adequately respond in DA. However, the GlotLID results indicate that they are not always using the correct dialect. This is further shown in \autoref{tab:dialect-steering-by-country}, where the models' responses do not always follow the intended country-level dialect, as indicated by the varying GlotLID accuracy scores. Refer to Appendix~\ref{sec:dialect_steering_appendix} for further discussion.
\pagebreak
{\setlength{\tabcolsep}{0.2em}
\renewcommand{\arraystretch}{1.5}
\begin{table}[t]
\centering
\small
\begin{tabular}{l|cc|cc}
\toprule
\textbf{Model} & \multicolumn{2}{c|}{\textbf{Target: MSA}} & \multicolumn{2}{c}{\textbf{Target: Dialect}} \\
 & \textbf{Judge} & \textbf{Acc\textsubscript{Dialect}} & \textbf{Judge} & \textbf{Acc\textsubscript{Dialect}} \\
\midrule
\rowcolor{headerpink}\multicolumn{5}{c}{\textbf{Proprietary Models (0-shot)}} \\
Gemini-2.5-pro & 0.895 & \underline{0.757} & 0.924 & \underline{\textbf{0.505}} \\
GPT-5 & \underline{\textbf{0.931}} & 0.718 & \underline{\textbf{0.956}} & 0.454 \\
\midrule
\rowcolor{headergreen}\multicolumn{5}{c}{\textbf{Arabic-centric Models (0-shot)}} \\
Jais-7B-chat & 0.782 & 0.827 & 0.599 & 0.022 \\
ALLaM-7B & \underline{{0.894}} & 0.827 & \underline{{0.8168}} & \underline{{0.3625}} \\
Cohere-Arabic-7B & 0.843 & 0.794 & 0.716 & 0.186 \\
Jais-2.8B-chat & 0.860 & 0.789 & 0.730 & 0.208 \\
Fanar-1.9B & 0.792 & 0.776 & 0.621 & 0.038 \\
SILMA-9B-it & 0.768 & 0.666 & 0.591 & 0.078 \\
Hala-9B & 0.829 & \underline{\textbf{0.874}} & 0.612 & 0.016 \\
\midrule
\rowcolor{headergold}\multicolumn{5}{c}{\textbf{Multilingual Models (0-shot)}} \\
LLaMA-3.1-8B-it & 0.739 & 0.785 & 0.516 & 0.064 \\
Qwen3-8B & 0.549 & \underline{{0.827}} & 0.430 & 0.041 \\
Gemma-2-9B-it & \underline{{0.793}} & 0.780 & \underline{{0.604}} & \underline{{0.164}} \\
\midrule
\rowcolor{headerblue}\multicolumn{5}{c}{\textbf{Multilingual Models (SFT)}} \\
LLaMA-3.1-8B-it & 0.750 & 0.700 & 0.514 & 0.127 \\
Qwen3-8B & 0.800 & 0.744 & 0.536 & 0.077\\
Gemma-2-9B-it & \underline{{0.825}}& \underline{{0.769}} & \underline{{0.604}} & \underline{{0.170}} \\
\bottomrule
\end{tabular}
\caption{\textbf{Task 3 - Dialect Steering}'s results averaged across all dialogues. Zero-shot performance is reported for all models, and supervised fine-tuning (SFT) results are reported for multilingual ones. Judge scores $\in$ [0,1]. Acc\textsubscript{Dialect} reports GlotLID's strict ISO-code accuracy. \underline{Underline} indicates the best score for each model category, and \textbf{bold} indicates the overall best.
}
\label{tab:dialect-steering-zero-shot-plus-sft}
\end{table}
}

\paragraph{Supervised steering shifts quality and dialect identity.:} Fine-tuning improves the quality of the multilingual models' responses as indicated by the judge's scores, especially for MSA. However, the models' ability to reply in MSA decreases, as indicated by the lower dialect accuracy scores. When the target is generating outputs in a specific dialect, SFT improves the models' ability to choose the right dialect in aggregate, yet the gap is still significant. However, the gains vary from one country-level dialect to another (see \autoref{tab:dialect-steering-by-country-sft}). For instance, Moroccan responses seem to benefit more from SFT than other dialects. The excerpts in \autoref{fig:steering_excerpt} suggest that SFT could be pushing the model to generate distinctive cues of each dialect (bolded) rather than pan-Arabic colloquialism. Overall, this reflects a trade-off between output quality and dialect control under fine-tuning.

\begin{figure}
    \centering
    \begin{quote}
    \footnotesize
    \textbf{Gemma-2-9B-it (UAE, zero-shot):} \newline \begin{arab}كل عام وانت بالف خير يا هزيم، وربي يبارك فيك ويحفظك.\end{arab}\newline
    (Wishing you well every year, Hazim. May God bless you and protect you.)
    \par
    \textbf{Gemma-2-9B-it (UAE, SFT):} \newline \begin{arab}لدي، نقول كل عام وأنت بخير \textbf{عشان نشوف} الفرحة في عيون الناس.
    \end{arab}\newline
    (Alright, let’s say ‘happy every year’ and see what happens.)
    \par
    \textbf{Gemma-2-9B-it (Morocco, zero-shot):} \newline \begin{arab}والليلة \textbf{هادفنا} نعمل السُّمن\end{arab}\newline
    (And tonight, our goal is to make ghee.)
    \par
    \textbf{Gemma-2-9B-it (Morocco, SFT):} \newline \begin{arab}\textbf{بصح، باش نستاهل} الشهر الفضيل، نعمل السلو والسحلب \textbf{باش نحتفظو} به حتى رمضان.\end{arab}\newline
    (Right, so that we’re ready for the blessed month, we make sellou and sahlab and keep them until Ramadan.)
    \end{quote}
    \caption{The impact of SFT on the generated responses of Gemma-2 (a multilingual LLM), for the Dialect Steering task.}
    \label{fig:steering_excerpt}

\end{figure}

%% file: sections/5_conclusion.tex
\vspace{6pt}
\section{Conclusion and Future Work}

We introduce \textbf{ArabCulture-Dialogue}, the first culturally grounded conversational dataset covering 13 Arabic-speaking countries, spanning both MSA and corresponding dialects across 12 everyday domains and 54 fine-grained subtopics, with a total of 343,804 words. We use it to evaluate three tasks: (i) multiple-choice cultural reasoning, (ii) translation between MSA and dialects, and (iii) dialect-steered generation. The dataset supports evaluation in both controlled and open-ended settings, capturing variation across countries and contexts.

Our results show that while proprietary models perform strongly on cultural reasoning MCQs, open-weight models, particularly at the 7B scale, often struggle, in some cases approaching random guessing; similar weaknesses appear in dialect translation and dialect steering, where all model types exhibit limited dialectal competence. These findings expose substantial gaps in current open-weight LLMs’ ability to model culturally grounded, dialect-rich Arabic, especially in conversational settings. These limitations point to promising directions for future work, including dialect-aware pretraining and instruction tuning, expanding coverage to additional Arab countries and dialects, and developing models that better integrate cultural knowledge in conversation.
\pagebreak

%% file: sections/6_limitations.tex
\newpage
\section*{Limitations}
While our dialogue data provides translations into 13 different country-level dialects covering the different regions of the Arab world, it still does not cover all Arab countries. Additionally, we acknowledge the interspeaker dialectal variation that exists within each Arabic-speaking country.

Despite the efforts to ensure a high-quality translation of the MSA dialogues, the translators were inevitably impacted by the MSA dialogues' style (e.g., syntax). Hence, signs of translationese can still be noticed in some translations, a limitation previously reported in \citet{bouamor-etal-2014-multidialectal, bouamor-etal-2018-madar}.

\section*{Ethics and Broader Impact}

This benchmark is designed to evaluate LLMs’ cultural reasoning abilities across Arabic-speaking countries in both MSA and regional dialects. Beyond evaluation, the dataset can also be used for training models to improve their understanding of culturally grounded Arabic language use. However, several considerations must be acknowledged. Cultural practices often overlap across countries, and not all instances in the dataset represent strictly country-specific culture; such distinctions are explicitly annotated. Additionally, the benchmark does not aim to capture the full cultural diversity of the Arab world, as it covers 13 of the 22 Arab countries and therefore represents only a subset of Arab cultural practices. These limitations should be taken into account when interpreting results or deploying models trained or evaluated using this dataset.

Additionally, annotators provided agreement to participate in this initiative and were informed that the data would be used for benchmarking purposes. Since their work involved refining existing content rather than creating data from scratch, no personally identifiable information is included in the dataset.

\section*{Acknowledgments}

We thank Badr M. Abdullah and Fadhl Eryani for their help in checking the quality of some Yemeni translations.
This research is partially supported by the Dubai Research Development and Innovation (RDI) Grant\footnote{\url{https://dubairdi.ae/}} and the MBZUAI supercomputing cluster.

%% file: sections/7_appendix.tex
\newpage
\appendix
\setcounter{table}{0}
\setcounter{figure}{0}
\renewcommand{\thetable}{\Alph{section}\arabic{table}}
\renewcommand{\thefigure}{\Alph{section}\arabic{figure}}

\section{Data Statistics Details}
\label{sec:appendix-data-stats-region-topic}

\autoref{tab:appendix-data-stats-all} reports on the number of dialogues, with the accompanying statistics for each country, independently. \autoref{tab:appendix-data-stats-topic} further presents dataset statistics broken down by topic. \textbf{ArabCulture-Dialogue} covers 12 topics that reflect culturally grounded aspects of everyday Arabic life.

\setlength{\tabcolsep}{0.1em}
\begin{table*}[!h]
\centering
\small
\resizebox{\textwidth}{!}{%
\begin{tabular}{l | cccc | cc | cccc | ccc}
\toprule
\textbf{Metric} 
& \textbf{Algeria} & \textbf{Libya} & \textbf{Morocco} & \textbf{Tunisia}
& \textbf{Egypt} & \textbf{Sudan}
& \textbf{Jordan} & \textbf{Lebanon} & \textbf{Palestine} & \textbf{Syria}
& \textbf{KSA} & \textbf{UAE} & \textbf{Yemen} \\
\midrule

\multicolumn{14}{c}{\textbf{General Dialogue Statistics}} \\
\#dialogues 
& 271 & 239 & 276 & 250
& 265 & 256
& 290 & 255 & 273 & 279
& 261 & 283 & 273 \\
\#country Specific 
& 81 & 100 & 103 & 164
& 197 & 144
& 17 & 99 & 29 & 47
& 98 & 105 & 206 \\
\midrule

\multicolumn{14}{c}{\textbf{Modern Standard Arabic (MSA) Data}} \\
avg. words per dial. 
& 51.66 & 52.26 & 50.35 & 50.12
& 50.46 & 51.03
& 50.79 & 51.39 & 49.90 & 49.36
& 50.26 & 48.95 & 51.15 \\
avg. utt per dial. 
& 6.06 & 6.05 & 6.02 & 6.04
& 6.04 & 6.03
& 6.10 & 6.05 & 6.24 & 5.81
& 6.10 & 6.08 & 6.09 \\
avg. words per utt. 
& 7.53 & 7.64 & 7.36 & 7.31
& 7.36 & 7.47
& 7.34 & 7.50 & 7.04 & 7.46
& 7.26 & 7.06 & 7.42 \\
\#words 
& 13,999 & 12,491 & 13,896 & 12,531
& 13,372 & 13,064
& 14,728 & 13,104 & 13,624 & 13,772
& 13,118 & 13,853 & 13,963 \\
\#unique words 
& 4,309 & 4,133 & 4,181 & 3,929
& 4,246 & 3,983
& 4,125 & 4,415 & 4,259 & 4,518
& 4,112 & 4,565 & 4,251 \\
\midrule

\multicolumn{14}{c}{\textbf{Dialect Data}} \\
avg. words per dial. 
& 48.46 & 51.08 & 50.96 & 50.24
& 49.29 & 50.79
& 50.21 & 47.53 & 45.61 & 43.42
& 47.02 & 45.84 & 50.42 \\
avg. utt. per dial. 
& 6.06 & 6.05 & 6.02 & 6.04
& 6.04 & 6.03
& 6.10 & 6.05 & 6.24 & 5.81
& 6.10 & 6.08 & 6.09 \\
avg. words per utt. 
& 7.01 & 7.41 & 7.47 & 7.34
& 7.17 & 7.43
& 7.24 & 6.88 & 6.29 & 6.44
& 6.72 & 6.53 & 7.30 \\
\#words 
& 13,134 & 12,208 & 14,065 & 12,561
& 13,063 & 13,003
& 14,560 & 12,121 & 12,451 & 12,115
& 12,271 & 12,972 & 13,765 \\
\#unique words 
& 4,259 & 4,204 & 4,355 & 3,846
& 4,100 & 4,071
& 4,249 & 4,631 & 4,674 & 4,314
& 4,111 & 4,428 & 4,325 \\
\bottomrule
\end{tabular}%
}
\caption{Detailed dataset statistics, split by country.}
\label{tab:appendix-data-stats-all}
\end{table*}

\setlength{\tabcolsep}{0.15em}
\begin{table*}[t]
\centering
\small
\resizebox{\textwidth}{!}{%
\begin{tabular}{l | cccccccccccc}
\toprule
\textbf{Metric} & \textbf{Food} & \textbf{Daily} & \textbf{Holiday} & \textbf{Habits} & \textbf{Weddings} & \textbf{Death} & \textbf{Art} & \textbf{Games} & \textbf{Idiom} & \textbf{Parenting} & \textbf{Family} & \textbf{Agriculture} \\
\midrule
\multicolumn{13}{c}{\textbf{General Dialogue Statistics (By Topic)}} \\
\#dialogues & 724 & 520 & 461 & 342 & 279 & 239 & 234 & 223 & 118 & 116 & 108 & 107 \\
\#country Specific & 357 & 158 & 175 & 85 & 145 & 42 & 166 & 85 & 82 & 23 & 21 & 51 \\
\midrule
\multicolumn{13}{c}{\textbf{Modern Standard Arabic (MSA) Data}} \\
avg. words per dial. & 49.56 & 50.09 & 50.84 & 51.65 & 49.39 & 51.19 & 50.56 & 50.50 & 53.91 & 50.90 & 50.71 & 52.64 \\
avg. utt per dial. & 6.13 & 6.03 & 6.03 & 6.04 & 6.04 & 5.99 & 5.98 & 6.01 & 6.13 & 6.03 & 6.00 & 6.31 \\
avg. words per utt. & 7.11 & 7.32 & 7.44 & 7.56 & 7.18 & 7.54 & 7.45 & 7.41 & 7.79 & 7.45 & 7.45 & 7.40 \\
\#words & 35,883 & 26,047 & 23,437 & 17,665 & 13,780 & 12,234 & 11,832 & 11,262 & 6,361 & 5,904 & 5,477 & 5,633 \\
\#unique words & 6,138 & 5,899 & 4,845 & 5,149 & 3,377 & 2,888 & 3,168 & 3,061 & 2,238 & 2,374 & 2,244 & 1,933 \\
\midrule
\multicolumn{13}{c}{\textbf{Dialect Data}} \\
avg. words per dial. & 47.22 & 47.97 & 48.51 & 49.75 & 47.83 & 48.97 & 48.84 & 48.84 & 52.84 & 48.59 & 48.09 & 49.98 \\
avg. utt. per dial. & 6.13 & 6.03 & 6.03 & 6.04 & 6.04 & 5.99 & 5.98 & 6.01 & 6.13 & 6.03 & 6.00 & 6.31 \\
avg. words per utt. & 6.70 & 6.96 & 7.05 & 7.24 & 6.92 & 7.17 & 7.16 & 7.13 & 7.60 & 7.08 & 7.02 & 6.99 \\
\#words & 34,186 & 24,945 & 22,361 & 17,014 & 13,345 & 11,703 & 11,429 & 10,892 & 6,235 & 5,637 & 5,194 & 5,348 \\
\#unique words & 8,703 & 7,632 & 6,509 & 6,309 & 4,286 & 3,792 & 3,954 & 3,966 & 2,616 & 2,789 & 2,564 & 2,316 \\
\bottomrule
\end{tabular}%
}
\caption{Detailed dataset statistics by topic.
}
\label{tab:appendix-data-stats-topic}
\end{table*}

\section{Annotator Requirements}
\label{sec:appendix-annotator}

In our study, we specified several requirements for the annotators: they had to be fluent in both MSA and their respective dialects, have lived in their country for more than 10 years, and possess a strong understanding of local cultural norms. Additionally, annotators were required to have completed at least a high-school level education. Based on these criteria, we recruited 26 annotators who met all requirements. Before starting the annotation process, we conducted workshops to explain the task in detail and provided a comprehensive guideline document. To ensure full understanding, we asked the annotators to complete a set of sample annotations before the main phase. The authors reviewed these samples to verify quality and consistency before proceeding with the full annotation process. We also further ensured the quality of the annotations by conducting a double quality-check procedure, as mentioned in Section~\ref{sec:dataset-construction}. This process helped maintain consistency and reliability across annotators and countries, ensuring high-quality and culturally accurate annotations. 

\section{Task-specific Prompts}
\label{sec:appendix-task-prompts}

\begin{figure}[!h]
  \centering
\todo[inline, color=teal!20]
{\small You are a professional Arab able to reason about the Arab culture.

\vspace{6pt}
Rules:

- Output only the OPTIONS [A/B/C]
    
- Do not add explanations, comments, or quotation marks. Only the option label [A/B/C]

\vspace{6pt}
You are tasked with selecting the most culturally appropriate option based on the context provided below.

\vspace{6pt}
Location: \{country\}, \{region\}

Conversation:

\{dialogue\}

\vspace{6pt}
Consider the cultural nuances of the specified location and choose the most suitable next utterance! Give the option label only [A/B/C]

\vspace{6pt}
Options:
\{choices\}}
  \caption{\textbf{Task 1 - MCQ Evaluation}'s prompt.}
  \label{fig:appendix-prompt-mcq}
\end{figure}

All prompts are written in English, following the findings of \citealp{sadallah-etal-2025-commonsense}, which show that current Arabic-specific and multilingual models achieve better performance when prompted in English rather than in MSA. \autoref{fig:appendix-prompt-mcq}, \autoref{fig:appendix-prompt-mt}, and \autoref{fig:appendix-promtp-steering} present the prompts used for the MCQ Evaluation, Dialect Translation, and Dialect Steering tasks, respectively.

\begin{figure}[!h]
  \centering

  \begin{subfigure}[t]{\linewidth}
    \centering
    \begin{tcolorbox}[colback=gray!5, colframe=black!40]
\small
You are a professional Arabic translation system specialized in
Modern Standard Arabic (MSA) and regional Arabic dialects.

Rules:
\begin{itemize}[noitemsep, topsep=2pt, leftmargin=*]
    \item Output only the translated text in Arabic.
    \item Do not add explanations, comments, labels, or quotation marks.
    \item Preserve meaning, tone, and level of formality.
    \item Do not change names, numbers, or entities.
    \item If the input already matches the target variety, return it unchanged.
\end{itemize}
    \end{tcolorbox}
    \caption{Unified system prompt.}
  \end{subfigure}

  \vspace{8pt}

  \begin{subfigure}[t]{\linewidth}
    \centering
    \begin{tcolorbox}[colback=blue!5, colframe=black!40]
\small
Translate the following text from Modern Standard Arabic (MSA)
into the \{region\} dialect of \{country\}.

Text:
\{record\}
    \end{tcolorbox}
    \caption{User prompt for MSA-to-Dialect translation.}
  \end{subfigure}

  \vspace{8pt}

  \begin{subfigure}[t]{\linewidth}
    \centering
    \begin{tcolorbox}[colback=green!5, colframe=black!40]
\small
Translate the following text from the \{region\} dialect of \{country\}
into Modern Standard Arabic (MSA).

Text:
\{record\}
    \end{tcolorbox}
    \caption{User prompt for Dialect-to-MSA translation.}
  \end{subfigure}

  \caption{\textbf{Task 2 - Dialect Translation}'s prompt. A unified system prompt defines the translation role and constraints, while direction-specific user prompts specify the translation task and input text.}
  \label{fig:appendix-prompt-mt}
\end{figure}

\begin{figure}[!h]
\begin{subfigure}[t]{\linewidth}
\centering
\todo[inline, color=magenta!20]
{\small You are a helpful assistant who writes only in Modern Standard Arabic (MSA). Continue the dialogue with a single natural utterance and avoid extra explanations.

\vspace{6pt}
Continue the following dialogue in Modern Standard Arabic (MSA): \{record\}
}
  \caption{Generating MSA outputs.}
  \label{fig:appendix-prompt-completion}
\end{subfigure}%
\hfill
\begin{subfigure}[t]{\linewidth}
  \centering
\todo[inline, color=magenta!20]
{\small You are a helpful assistant who writes in the \{dialect\_name\} dialect (code \{code\}). Continue the dialogue with one natural utterance in that dialect without translation or commentary.

\vspace{6pt}
Continue the following dialogue in \{dialect\_name\} while keeping the conversational tone: \{record\}}
  \caption{Generating Dialectal outputs.}
  \label{fig:appendix-prompt-steer}
\end{subfigure}
\caption{\textbf{Task 3 - Dialect Steering}'s prompt.}
\label{fig:appendix-promtp-steering}
\end{figure}

\section{Task 1 - MCQ Evaluation Details}
\label{sec:appendix-mcq-eval-details}

\begin{table*}[!h]
\begin{subtable}[t]{0.98\textwidth}
\centering
\small
\begin{tabular}{l | ccc | ccc | ccc}
\toprule
\multirow{2}{*}{\textbf{Model}} & \multicolumn{3}{c|}{\textbf{Context: None}} & \multicolumn{3}{c|}{\textbf{Context: Reg.}} & \multicolumn{3}{c}{\textbf{Context: Reg. + Cou.}} \\
\cline{2-10}
\\[-0.8em]
 & \textbf{Acc} & \textbf{CS} & \textbf{\textasciitilde CS} & \textbf{Acc} & \textbf{CS} & \textbf{\textasciitilde CS} & \textbf{Acc} & \textbf{CS} & \textbf{\textasciitilde CS} \\
\midrule
Random & 0.333 & 0.333 & 0.333 & 0.333 & 0.333 & 0.333 & 0.333 & 0.333 & 0.333 \\
\midrule
\rowcolor{headerpink} \multicolumn{10}{c}{\textbf{Proprietary Models}} \\
Gemini-2.5-pro & 0.942 & 0.920 & 0.954 & 0.943 & 0.922 & 0.958 & 0.950 & 0.933 & 0.961 \\
GPT-5 & 0.943 & 0.908 & 0.966 & 0.946 & 0.912 & 0.968 & 0.953 & 0.927 & 0.971 \\
\midrule
\rowcolor{headergreen} \multicolumn{10}{c}{\textbf{Arabic Centric Models}} \\
Jais-7B-chat & 0.554 & 0.504 & 0.589 & 0.532 & 0.488 & 0.563 & 0.531 & 0.492 & 0.558 \\
ALLaM-7B-Instruct & 0.418 & 0.387 & 0.437 & 0.474 & 0.437 & 0.499 & 0.506 & 0.465 & 0.533 \\
Cohere-Arabic-7B & 0.355 & 0.366 & 0.348 & 0.340 & 0.351 & 0.334 & 0.375 & 0.385 & 0.368 \\
Jais-2 8B-Chat & 0.740 & 0.689 & 0.774 & 0.754 & 0.701 & 0.789 & 0.766 & 0.704 & 0.808 \\
Fanar-1-9B & 0.391 & 0.373 & 0.403 & 0.444 & 0.410 & 0.467 & 0.618 & 0.553 & 0.662 \\
SILMA-9B-Instruct & \underline{\textbf{0.783}} & \underline{\textbf{0.715}} & \underline{\textbf{0.829}} & 0.780 & 0.714 & 0.824 & 0.784 & 0.724 & 0.825 \\
Hala-9B & 0.779 & 0.711 & 0.826 & \underline{\textbf{0.810}} & \underline{\textbf{0.739}} & \underline{\textbf{0.858}} & \underline{\textbf{0.820}} & \underline{\textbf{0.751}} & \underline{\textbf{0.866}} \\
\midrule
\rowcolor{headergold} \multicolumn{10}{c}{\textbf{Multilingual Models}} \\
Llama-3.1-8B-it & 0.456 & 0.424 & 0.478 & 0.463 & 0.430 & 0.486 & 0.473 & 0.441 & 0.495 \\
Qwen-3-8B & 0.354 & 0.350 & 0.357 & 0.370 & 0.361 & 0.375 & 0.380 & 0.368 & 0.387 \\
Gemma-2-9B-it & \underline{0.671} & \underline{0.619} & \underline{0.707} & \underline{0.702} & \underline{0.648} & \underline{0.739} & \underline{0.710} & \underline{0.661} & \underline{0.744} \\
\bottomrule
\end{tabular}
\caption{MSA dialogues.}
\label{tab:appendix-msa-mcq-result}
\end{subtable}
\hfill
\begin{subtable}[t]{0.98\textwidth}
\centering
\small
\begin{tabular}{l | ccc | ccc | ccc}
\toprule
\multirow{2}{*}{\textbf{Model}} & \multicolumn{3}{c|}{\textbf{Context: None}} & \multicolumn{3}{c|}{\textbf{Context: Reg.}} & \multicolumn{3}{c}{\textbf{Context: Reg. + Cou.}} \\
\cline{2-10}
\\[-0.8em]
 & \textbf{Acc} & \textbf{CS} & \textbf{\textasciitilde CS} & \textbf{Acc} & \textbf{CS} & \textbf{\textasciitilde CS} & \textbf{Acc} & \textbf{CS} & \textbf{\textasciitilde CS} \\
\midrule
Random & 0.333 & 0.333 & 0.333 & 0.333 & 0.333 & 0.333 & 0.333 & 0.333 & 0.333 \\
\midrule
\rowcolor{headerpink} \multicolumn{10}{c}{\textbf{Proprietary Models}} \\
Gemini-2.5-pro & 0.939 & 0.917 & 0.956 & 0.939 & 0.912 & 0.960 & 0.939 & 0.924 & 0.950 \\
GPT-5 & 0.948 & 0.924 & 0.964 & 0.943 & 0.915 & 0.962 & 0.948 & 0.923 & 0.965 \\
\midrule
\rowcolor{headergreen} \multicolumn{10}{c}{\textbf{Arabic Centric Models}} \\
Jais-7B-chat & 0.498 & 0.460 & 0.524 & 0.471 & 0.440 & 0.491 & 0.469 & 0.440 & 0.489 \\
ALLaM-7B-Instruct & 0.398 & 0.374 & 0.413 & 0.435 & 0.411 & 0.450 & 0.471 & 0.440 & 0.491 \\
Cohere-Arabic-7B & 0.342 & 0.351 & 0.337 & 0.338 & 0.345 & 0.334 & 0.357 & 0.361 & 0.355 \\
Jais-2-8B-Chat & 0.710 & 0.650 & 0.751 & 0.715 & 0.647 & 0.761 & 0.731 & 0.667 & 0.773 \\
Fanar-1-9B & 0.350 & 0.354 & 0.348 & 0.379 & 0.369 & 0.386 & 0.534 & 0.486 & 0.566 \\
SILMA-9B-Instruct & 0.716 & \underline{\textbf{0.663}} & 0.751 & 0.715 & 0.664 & 0.749 & 0.714 & 0.665 & 0.747 \\
Hala-9B & \underline{\textbf{0.733}} & 0.660 & \underline{\textbf{0.782}} & \underline{\textbf{0.750}} & \underline{\textbf{0.676}} & \underline{\textbf{0.800}} & \underline{\textbf{0.763}} & \underline{\textbf{0.692}} & \underline{\textbf{0.810}} \\
\midrule
\rowcolor{headergold} \multicolumn{10}{c}{\textbf{Multilingual Models}} \\
Llama-3.1-8B-it & 0.412 & 0.406 & 0.416 & 0.427 & 0.415 & 0.436 & 0.430 & 0.422 & 0.435 \\
Qwen-3-8B & 0.354 & 0.346 & 0.359 & 0.366 & 0.351 & 0.376 & 0.368 & 0.355 & 0.377 \\
Gemma-2-9B-it & \underline{0.609} & \underline{0.558} & \underline{0.644} & \underline{0.641} & \underline{0.585} & \underline{0.679} & \underline{0.643} & \underline{0.582} & \underline{0.684} \\
\bottomrule
\end{tabular}
\caption{Dialectal dialogues.}
\label{tab:appendix-dialect-mcq-result}
\end{subtable}
\caption{Full results scores for \textbf{Task 1 - MCQ Cultural Reasoning}. Results are averaged and reported separately for country-specific (CS) and non-country-specific ($\sim$CS) dialogues. The best overall model is shown in \textbf{bold}, and the best within each category is \underline{underlined}.}
\label{tab:appendix-mcq-result}
\end{table*}

Tables~\ref{tab:appendix-msa-mcq-result} and~\ref{tab:appendix-dialect-mcq-result} report model performance on MSA and dialect data, respectively, for the dialogue-based multiple-choice cultural commonsense reasoning task in Arabic. Overall, performance on MSA dialogues is substantially higher than on dialect dialogues, and this trend is consistent across all Arabic-centric and multilingual models. In contrast, proprietary models exhibit comparable performance across both MSA and dialect settings. This robustness holds across all context configurations, including no geographic context, region-only context, and full context with both region and country information. 

These results suggest that dialectal variation introduces additional challenges that are not fully captured by MSA-based evaluation.

\subsection{MCQ Evaluation Analysis per Country}
\label{sec:appendix-mcq-eval-country}

We further analyze the performance of the strongest multilingual model (Gemma-2-9B-Instruct) and the strongest Arabic-centric model (Hala-9B) across countries, regions, and topics. As shown in Tables~\ref{tab:appendix-msa-mcq-comp_country} and~\ref{tab:appendix-dialect-mcq-comp_country}, both models achieve relatively strong performance on dialogues from Jordan and Palestine, suggesting that the cultural cues in these countries may be easier to infer compared to others. In contrast, dialogues from Yemen and the UAE are consistently the most challenging.

At the regional level, North Africa is the most challenging region, with performance dropping to 0.663 on country-specific dialogues in the dialect setting. This highlights the greater complexity and diversity of its dialectal and cultural expressions, which often differ from norms and linguistic patterns typically represented in MSA.

\begin{table*}[!h]
\begin{subtable}[t]{0.48\textwidth}
\centering
\small
\begin{tabular}{l | cc | cc}
\toprule
\multirow{2}{*}{\textbf{Country}} & \multicolumn{2}{c|}{\textbf{Gemma-2-9B-it}} & \multicolumn{2}{c}{\textbf{Hala-9B}} \\
\cline{2-5}
\\[-0.8em]
 & \textbf{CS} & \textbf{\textasciitilde CS} & \textbf{CS} & \textbf{\textasciitilde CS} \\
\midrule
Algeria & 0.654 & \cellcolor{red!25}0.653 & 0.716 & 0.826 \\
Egypt & \cellcolor{green!20}0.751 & 0.706 & 0.802 & \cellcolor{red!25}0.721 \\
Jordan & \cellcolor{green!20}0.882 & 0.777 & \cellcolor{green!20}0.882 & 0.923 \\
KSA & 0.704 & 0.772 & 0.847 & 0.877 \\
Lebanon & 0.636 & \cellcolor{red!25}0.635 & 0.758 & 0.821 \\
Libya & 0.730 & 0.727 & 0.790 & 0.878 \\
Morocco & 0.728 & \cellcolor{green!20}0.809 & 0.854 & \cellcolor{green!20}0.948 \\
Palestine & 0.690 & \cellcolor{green!20}0.783 & \cellcolor{green!20}0.897 & 0.869 \\
Sudan & 0.611 & 0.750 & 0.743 & \cellcolor{green!20}0.929 \\
Syria & 0.660 & 0.759 & 0.723 & 0.836 \\
Tunisia & 0.628 & 0.756 & 0.689 & 0.861 \\
UAE & \cellcolor{red!25}0.552 & 0.753 & \cellcolor{red!25}0.657 & 0.865 \\
Yemen & \cellcolor{red!25}0.597 & 0.723 & \cellcolor{red!25}0.675 & \cellcolor{red!25}0.723 \\
\midrule
\textbf{Gulf} & \cellcolor{red!25}0.611 & \cellcolor{green!20}0.756 & \cellcolor{red!25}0.712 & \cellcolor{red!25}0.847 \\
\textbf{Levant} & 0.672 & 0.749 & \cellcolor{green!20}0.781 & 0.869 \\
\textbf{Nile River} & \cellcolor{green!20}0.692 & 0.733 & 0.777 & 0.850 \\
\textbf{N. Africa} & 0.679 & \cellcolor{red!25}0.731 & 0.755 & \cellcolor{green!20}0.879 \\
\bottomrule
\end{tabular}
\caption{MSA dialogues.}
\label{tab:appendix-msa-mcq-comp_country}
\end{subtable}
\hfill
\begin{subtable}[t]{0.48\textwidth}
\centering
\small
\begin{tabular}{l | cc | cc}
\toprule
\multirow{2}{*}{\textbf{Country}} & \multicolumn{2}{c|}{\textbf{Gemma-2-9B-it}} & \multicolumn{2}{c}{\textbf{Hala-9B}} \\
\cline{2-5}
\\[-0.8em]
 & \textbf{CS} & \textbf{\textasciitilde CS} & \textbf{CS} & \textbf{\textasciitilde CS} \\
\midrule
Algeria & \cellcolor{red!25}0.506 & \cellcolor{red!25}0.600 & \cellcolor{red!25}0.605 & 0.800 \\
Egypt & 0.650 & 0.706 & 0.782 & \cellcolor{red!25}0.750 \\
Jordan & \cellcolor{green!20}0.824 & \cellcolor{green!20}0.747 & 0.824 & \cellcolor{green!20}0.879 \\
KSA & 0.684 & \cellcolor{green!20}0.735 & \cellcolor{green!20}0.827 & \cellcolor{green!20}0.870 \\
Lebanon & 0.596 & \cellcolor{red!25}0.635 & 0.667 & 0.776 \\
Libya & 0.710 & 0.655 & 0.780 & 0.755 \\
Morocco & \cellcolor{red!25}0.485 & 0.665 & 0.738 & 0.867 \\
Palestine & \cellcolor{green!20}0.724 & 0.725 & \cellcolor{green!20}0.828 & 0.836 \\
Sudan & 0.528 & 0.643 & 0.660 & 0.804 \\
Syria & 0.574 & 0.659 & 0.702 & 0.759 \\
Tunisia & 0.518 & 0.651 & \cellcolor{red!25}0.573 & 0.756 \\
UAE & 0.533 & 0.713 & 0.667 & 0.826 \\
Yemen & 0.553 & 0.723 & 0.621 & \cellcolor{red!25}0.631 \\
\midrule
\textbf{Gulf} & 0.579 & \cellcolor{green!20}0.723 & 0.682 & 0.812 \\
\textbf{Levant} & \cellcolor{green!20}0.630 & 0.699 & 0.714 & \cellcolor{green!20}0.819 \\
\textbf{Nile River} & 0.598 & 0.667 & \cellcolor{green!20}0.730 & \cellcolor{red!25}0.783 \\
\textbf{N. Africa} & \cellcolor{red!25}0.551 & \cellcolor{red!25}0.639 & \cellcolor{red!25}0.663 & 0.803 \\
\bottomrule
\end{tabular}
\caption{Dialectal dialogues.}
\label{tab:appendix-dialect-mcq-comp_country}
\end{subtable}
\caption{Detailed scores for \textbf{Task 1 - MCQ Cultural Reasoning}, split by country and region. Green highlights indicate the two highest-performing topics, while red highlights indicate the two lowest-performing topics.}
\end{table*}

\subsection{MCQ Evaluation Analysis per Topic}
\label{sec:appendix-mcq-eval-topic}

Tables~\ref{tab:appendix-msa-mcq-comp_topic} and~\ref{tab:appendix-dialect-mcq-comp_topic} present the performance of Gemma-2-9B and Hala-9B on country-specific and non-country-specific dialogues, grouped by topic. As observed, performance on country-specific dialogues is consistently lower than on non-country-specific dialogues, in both MSA and dialect settings. In addition, the easiest topics for the models are `agriculture' and `family relationships', while the most challenging topics are `death' and `food'. This indicates that topics involving more context-dependent or sensitive cultural norms are harder for models to reason about.

\begin{table*}[!h]
\begin{subtable}[t]{0.48\textwidth}
\centering
\small
\begin{tabular}{l | cc | cc}
\toprule
\multirow{2}{*}{\textbf{Topic}} & \multicolumn{2}{c|}{\textbf{Gemma-2-9B}} & \multicolumn{2}{c}{\textbf{Hala-9B}} \\
\cline{2-5}
\\[-0.8em]
 & \textbf{CS} & \textbf{\textasciitilde CS} & \textbf{CS} & \textbf{\textasciitilde CS} \\
\midrule
Agriculture & \cellcolor{green!20}0.765 & 0.839 & 0.765 & \cellcolor{green!20}0.946 \\
Art & 0.669 & 0.779 & 0.789 & 0.882 \\
Daily Act. & \cellcolor{red!25}0.614 & 0.746 & \cellcolor{red!25}0.715 & 0.890 \\
Death & \cellcolor{red!25}0.595 & \cellcolor{red!25}0.645 & 0.762 & 0.878 \\
Family Rel. & 0.667 & \cellcolor{green!20}0.839 & \cellcolor{green!20}0.857 & 0.931 \\
Food & 0.644 & 0.701 & \cellcolor{red!25}0.703 & \cellcolor{red!25}0.775 \\
Habits & 0.682 & 0.821 & 0.788 & 0.883 \\
Holiday Act. & 0.646 & 0.717 & 0.737 & 0.871 \\
Idioms & 0.659 & \cellcolor{red!25}0.583 & 0.744 & \cellcolor{red!25}0.806 \\
Parenting & \cellcolor{green!20}0.826 & \cellcolor{green!20}0.849 & 0.783 & \cellcolor{green!20}0.936 \\
Trd. Games & 0.659 & 0.703 & 0.800 & 0.891 \\
Wedding & 0.710 & 0.806 & \cellcolor{green!20}0.807 & 0.843 \\
\bottomrule
\end{tabular}
\caption{MSA dialogues.}
\label{tab:appendix-msa-mcq-comp_topic}
\end{subtable}
\hfill
\begin{subtable}[t]{0.48\textwidth}
\centering
\small
\begin{tabular}{l | cc | cc}
\toprule
\multirow{2}{*}{\textbf{Topic}} & \multicolumn{2}{c|}{\textbf{Gemma-2-9B}} & \multicolumn{2}{c}{\textbf{Hala-9B}} \\
\cline{2-5}
\\[-0.8em]
 & \textbf{CS} & \textbf{\textasciitilde CS} & \textbf{CS} & \textbf{\textasciitilde CS} \\
\midrule
Agriculture & 0.647 & \cellcolor{green!20}0.839 & 0.686 & \cellcolor{green!20}0.893 \\
Art & 0.602 & 0.721 & 0.735 & 0.838 \\
Daily Act. & 0.563 & 0.635 & 0.715 & 0.834 \\
Death & 0.548 & \cellcolor{red!25}0.609 & 0.738 & 0.812 \\
Family Rel. & \cellcolor{green!20}0.762 & 0.736 & \cellcolor{green!20}0.762 & \cellcolor{green!20}0.851 \\
Food & 0.555 & 0.651 & \cellcolor{red!25}0.611 & \cellcolor{red!25}0.720 \\
Habits & 0.635 & \cellcolor{green!20}0.763 & 0.694 & 0.817 \\
Holiday Act. & \cellcolor{red!25}0.543 & 0.710 & 0.674 & 0.843 \\
Idioms & 0.573 & \cellcolor{red!25}0.556 & 0.732 & 0.833 \\
Parenting & \cellcolor{green!20}0.696 & 0.731 & \cellcolor{red!25}0.609 & 0.839 \\
Trd. Games & \cellcolor{red!25}0.506 & 0.667 & 0.718 & \cellcolor{red!25}0.790 \\
Wedding & 0.655 & 0.716 & \cellcolor{green!20}0.793 & 0.821 \\
\bottomrule
\end{tabular}
\caption{Dialectal dialogues.}
\label{tab:appendix-dialect-mcq-comp_topic}
\end{subtable}
\caption{Detailed scores for \textbf{Task 1 - MCQ Cultural Reasoning}, split by topic. Green highlights indicate the two highest-performing topics, while red highlights indicate the two lowest-performing topics.}

\end{table*}

\section{Task 2 - Dialect Translation's trends}
\label{sec:dialect_translation_appendix}

Tables~\ref{tab:country_msa_dialect} and~\ref{tab:country_dialect_msa} show consistent performance differences across countries for both MSA-to-Dialect and Dialect-to-MSA translation, respectively. In the MSA-to-Dialect direction, several countries exhibit lower BLEU and LLM-as-a-judge register scores for non-proprietary models, with particularly large drops for Morocco and Tunisia. This pattern indicates greater difficulty in generating accurate country-specific dialect forms. Proprietary and Arabic-centric models reduce these gaps but do not fully eliminate them. 

The country-level differences are smaller across all metrics for Dialect-to-MSA translation. Similar patterns arise region-wise in Tables~\ref{tab:region_msa_dialect} and~\ref{tab:region_dialect_msa}. For MSA-to-Dialect translation, North Africa consistently yields lower scores across all model categories, while the other regions achieve higher and more stable performance. This contrast between North Africa and the remaining regions is consistent across evaluation metrics and model types. In contrast, Dialect-to-MSA results exhibit smaller regional differences, with all regions reaching similar levels of translation quality and exhibiting more uniform performance overall. This suggests generating dialect-specific outputs is more challenging than normalizing dialects into MSA.

Across both country-wise and region-wise settings, LLM-as-a-judge register scores display the largest variation between models. This indicates that dialectal correctness remains the primary challenge in Arabic dialect translation, even when semantic adequacy and fluency scores are relatively high. In contrast, BLEU and BERTScore tend to show smaller differences, suggesting that surface-level similarity and general meaning are easier to capture than precise dialectal usage. This gap highlights that models can produce fluent and semantically correct outputs while still failing to match the intended dialect.

\begin{table*}[!h]
\centering
\small
\setlength{\tabcolsep}{4.5pt}
\begin{tabular}{l | cccc | ccc | ccccc}
\toprule
\multirow{2}{*}{\textbf{Country}} &
\multicolumn{4}{c|}{\textbf{BLEU}} &
\multicolumn{3}{c|}{\textbf{BERTScore}} &
\multicolumn{5}{c}{\textbf{LLM-as-a-judge (1--5)}} \\
\cline{2-13}\\
& B1 & B2 & B3 & B4 & P & R & F1 &
Adeq. & Flu. & Reg. & Term. & Overall \\
\midrule

\rowcolor{headerpink}
\multicolumn{13}{c}{\textbf{Best Proprietary Model: GPT-5}} \\

Algeria   & 0.581 & 0.449 & 0.351 & 0.269 & 0.865 & 0.873 & 0.869 & 4.857 & 4.631 & 4.721 & 4.873 & 4.439 \\
Egypt     & 0.650 & 0.531 & 0.443 & 0.366 & 0.891 & 0.893 & 0.892 & 4.954 & 4.866 & 4.870 & 4.954 & 4.741 \\
Jordan    & 0.554 & 0.418 & 0.316 & 0.233 & 0.883 & 0.893 & 0.888 & 4.939 & 4.793 & 4.820 & 4.962 & 4.667 \\
KSA       & 0.639 & 0.518 & 0.425 & 0.342 & 0.894 & 0.898 & 0.896 & 4.902 & 4.689 & 4.638 & 4.940 & 4.511 \\
Lebanon   & 0.590 & 0.461 & 0.363 & 0.281 & 0.874 & 0.881 & 0.878 & 4.952 & 4.869 & 4.904 & 4.965 & 4.782 \\
Libya     & 0.584 & 0.448 & 0.349 & 0.265 & 0.877 & 0.883 & 0.880 & 4.865 & 4.707 & 4.665 & 4.884 & 4.498 \\
Morocco   & 0.590 & 0.459 & 0.357 & 0.269 & 0.882 & 0.882 & 0.882 & 4.871 & 4.577 & 4.726 & 4.851 & 4.436 \\
Palestine & 0.617 & 0.492 & 0.398 & 0.313 & 0.881 & 0.880 & 0.880 & 4.947 & 4.793 & 4.781 & 4.963 & 4.638 \\
Sudan     & 0.602 & 0.467 & 0.367 & 0.283 & 0.889 & 0.889 & 0.889 & 4.913 & 4.596 & 4.587 & 4.939 & 4.430 \\
Syria     & 0.558 & 0.422 & 0.323 & 0.239 & 0.857 & 0.872 & 0.864 & 4.952 & 4.877 & 4.940 & 4.964 & 4.805 \\
Tunisia   & 0.490 & 0.339 & 0.235 & 0.158 & 0.837 & 0.851 & 0.844 & 4.884 & 4.716 & 4.764 & 4.924 & 4.502 \\
UAE       & 0.654 & 0.536 & 0.443 & 0.361 & 0.893 & 0.900 & 0.897 & 4.918 & 4.671 & 4.545 & 4.929 & 4.482 \\
Yemen     & 0.529 & 0.388 & 0.290 & 0.209 & 0.864 & 0.861 & 0.862 & 4.931 & 4.370 & 3.520 & 4.874 & 3.963 \\

\midrule
\rowcolor{headergreen}
\multicolumn{13}{c}{\textbf{Best Arabic-Centric Model: ALLaM-7B}} \\

Algeria   & 0.388 & 0.244 & 0.156 & 0.097 & 0.818 & 0.812 & 0.815 & 4.387 & 3.307 & 2.097 & 4.151 & 3.029 \\
Egypt     & 0.541 & 0.404 & 0.314 & 0.241 & 0.856 & 0.855 & 0.856 & 3.912 & 3.958 & 4.167 & 4.243 & 3.678 \\
Jordan    & 0.490 & 0.351 & 0.261 & 0.188 & 0.861 & 0.864 & 0.862 & 4.126 & 3.885 & 3.487 & 4.387 & 3.625 \\
KSA       & 0.554 & 0.418 & 0.322 & 0.244 & 0.866 & 0.866 & 0.866 & 4.183 & 4.068 & 3.349 & 4.523 & 3.694 \\
Lebanon   & 0.524 & 0.390 & 0.296 & 0.218 & 0.856 & 0.853 & 0.854 & 4.149 & 3.790 & 3.620 & 4.249 & 3.638 \\
Libya     & 0.545 & 0.423 & 0.337 & 0.262 & 0.858 & 0.856 & 0.857 & 4.377 & 3.623 & 2.512 & 4.237 & 3.261 \\
Morocco   & 0.380 & 0.237 & 0.145 & 0.082 & 0.799 & 0.780 & 0.789 & 4.202 & 2.968 & 2.057 & 3.887 & 2.835 \\
Palestine & 0.535 & 0.400 & 0.301 & 0.218 & 0.862 & 0.857 & 0.859 & 4.171 & 3.951 & 3.667 & 4.358 & 3.671 \\
Sudan     & 0.538 & 0.396 & 0.301 & 0.221 & 0.862 & 0.861 & 0.861 & 4.244 & 3.778 & 2.439 & 4.026 & 3.148 \\
Syria     & 0.512 & 0.376 & 0.277 & 0.197 & 0.852 & 0.859 & 0.856 & 4.132 & 3.896 & 3.749 & 4.295 & 3.685 \\
Tunisia   & 0.412 & 0.255 & 0.163 & 0.101 & 0.818 & 0.815 & 0.817 & 4.293 & 3.373 & 2.387 & 4.102 & 3.129 \\
UAE       & 0.596 & 0.470 & 0.377 & 0.296 & 0.877 & 0.878 & 0.877 & 4.094 & 3.890 & 3.188 & 4.255 & 3.596 \\
Yemen     & 0.480 & 0.345 & 0.254 & 0.181 & 0.843 & 0.838 & 0.841 & 4.358 & 3.976 & 2.386 & 4.398 & 3.260 \\

\midrule
\rowcolor{headergold}
\multicolumn{13}{c}{\textbf{Best Multilingual Model (0-shot): Qwen-3-8B}} \\

Algeria   & 0.334 & 0.195 & 0.111 & 0.064 & 0.800 & 0.794 & 0.797 & 3.353 & 2.635 & 1.402 & 3.143 & 2.197 \\
Egypt     & 0.384 & 0.235 & 0.146 & 0.086 & 0.814 & 0.809 & 0.811 & 3.272 & 2.490 & 1.707 & 3.318 & 2.339 \\
Jordan    & 0.400 & 0.253 & 0.160 & 0.100 & 0.833 & 0.833 & 0.833 & 3.548 & 2.835 & 1.556 & 3.513 & 2.444 \\
KSA       & 0.464 & 0.320 & 0.226 & 0.154 & 0.837 & 0.843 & 0.840 & 3.672 & 3.289 & 1.455 & 3.643 & 2.498 \\
Lebanon   & 0.387 & 0.249 & 0.159 & 0.099 & 0.815 & 0.809 & 0.812 & 3.424 & 2.638 & 1.594 & 3.415 & 2.389 \\
Libya     & 0.506 & 0.378 & 0.288 & 0.215 & 0.851 & 0.850 & 0.850 & 3.423 & 2.721 & 1.623 & 3.191 & 2.419 \\
Morocco   & 0.325 & 0.190 & 0.104 & 0.054 & 0.779 & 0.759 & 0.769 & 3.109 & 2.242 & 1.250 & 2.883 & 2.089 \\
Palestine & 0.395 & 0.253 & 0.166 & 0.105 & 0.819 & 0.812 & 0.816 & 3.557 & 2.837 & 1.524 & 3.541 & 2.439 \\
Sudan     & 0.464 & 0.322 & 0.229 & 0.157 & 0.838 & 0.837 & 0.837 & 3.296 & 2.783 & 1.439 & 3.239 & 2.278 \\
Syria     & 0.373 & 0.232 & 0.150 & 0.095 & 0.806 & 0.813 & 0.809 & 3.566 & 2.869 & 1.697 & 3.574 & 2.502 \\
Tunisia   & 0.332 & 0.183 & 0.100 & 0.055 & 0.797 & 0.793 & 0.795 & 3.227 & 2.444 & 1.436 & 2.951 & 2.191 \\
UAE       & 0.493 & 0.356 & 0.262 & 0.187 & 0.840 & 0.845 & 0.842 & 3.643 & 3.208 & 1.439 & 3.631 & 2.439 \\
Yemen     & 0.440 & 0.302 & 0.207 & 0.135 & 0.828 & 0.826 & 0.827 & 3.317 & 2.744 & 1.549 & 3.329 & 2.370 \\

\midrule
\rowcolor{headerblue}
\multicolumn{13}{c}{\textbf{Best Multilingual Model (SFT): Gemma-2-9B}} \\

Algeria   & 0.339 & 0.224 & 0.142 & 0.088 & 0.799 & 0.776 & 0.787 & 2.062 & 2.307 & 1.971 & 2.656 & 2.094 \\
Egypt     & 0.406 & 0.287 & 0.203 & 0.142 & 0.816 & 0.795 & 0.805 & 2.067 & 2.511 & 2.803 & 3.063 & 2.352 \\
Jordan    & 0.375 & 0.261 & 0.179 & 0.122 & 0.830 & 0.808 & 0.819 & 2.084 & 2.632 & 2.376 & 3.103 & 2.303 \\
KSA       & 0.450 & 0.333 & 0.247 & 0.180 & 0.836 & 0.815 & 0.825 & 2.089 & 2.728 & 2.396 & 3.098 & 2.311 \\
Lebanon   & 0.386 & 0.272 & 0.188 & 0.127 & 0.815 & 0.788 & 0.801 & 2.092 & 2.546 & 2.424 & 3.022 & 2.319 \\
Libya     & 0.457 & 0.357 & 0.282 & 0.217 & 0.833 & 0.809 & 0.820 & 2.051 & 2.535 & 1.944 & 2.884 & 2.154 \\
Morocco   & 0.328 & 0.218 & 0.139 & 0.085 & 0.796 & 0.767 & 0.781 & 2.016 & 2.129 & 1.960 & 2.468 & 2.004 \\
Palestine & 0.388 & 0.269 & 0.190 & 0.130 & 0.819 & 0.793 & 0.806 & 2.069 & 2.557 & 2.455 & 3.029 & 2.260 \\
Sudan     & 0.440 & 0.333 & 0.255 & 0.191 & 0.836 & 0.812 & 0.824 & 2.065 & 2.578 & 2.009 & 2.948 & 2.161 \\
Syria     & 0.367 & 0.244 & 0.165 & 0.107 & 0.811 & 0.797 & 0.804 & 2.064 & 2.542 & 2.518 & 3.068 & 2.307 \\
Tunisia   & 0.288 & 0.168 & 0.093 & 0.054 & 0.789 & 0.770 & 0.779 & 2.080 & 2.307 & 1.698 & 2.636 & 2.067 \\
UAE       & 0.443 & 0.331 & 0.249 & 0.182 & 0.834 & 0.814 & 0.824 & 2.098 & 2.569 & 1.945 & 2.996 & 2.184 \\
Yemen     & 0.383 & 0.273 & 0.193 & 0.134 & 0.818 & 0.793 & 0.805 & 2.085 & 2.752 & 1.854 & 3.224 & 2.207 \\

\bottomrule
\end{tabular}

\caption{Country-wise MSA-to-Dialect translation performance under \emph{Context: Country + Region}, reporting results for the best-performing model in each category: proprietary (GPT-5), multilingual (Qwen-3-8B), Arabic-centric (ALLaM-7B), and multilingual SFT (Gemma-2-9B). We report BLEU, BERTScore, and LLM-as-a-judge scores on a 1--5 scale.}
\label{tab:country_msa_dialect}
\end{table*}

\begin{table*}[!h]
\centering
\small
\setlength{\tabcolsep}{4.5pt}
\begin{tabular}{l | cccc | ccc | ccccc}
\toprule
\multirow{2}{*}{\textbf{Country}} &
\multicolumn{4}{c|}{\textbf{BLEU}} &
\multicolumn{3}{c|}{\textbf{BERTScore}} &
\multicolumn{5}{c}{\textbf{LLM-as-a-judge (1--5)}} \\
\cline{2-13}\\
& B1 & B2 & B3 & B4 & P & R & F1 &
Adeq. & Flu. & Reg. & Term. & Overall \\
\midrule

\rowcolor{headerpink}
\multicolumn{13}{c}{\textbf{Best Proprietary Model: GPT-5}} \\

Algeria   & 0.643 & 0.528 & 0.439 & 0.360 & 0.902 & 0.898 & 0.900 & 4.893 & 4.865 & 4.959 & 4.918 & 4.742 \\
Egypt     & 0.626 & 0.509 & 0.421 & 0.347 & 0.897 & 0.898 & 0.898 & 4.921 & 4.883 & 4.950 & 4.925 & 4.782 \\
Jordan    & 0.766 & 0.683 & 0.612 & 0.545 & 0.935 & 0.936 & 0.935 & 4.985 & 4.927 & 4.962 & 4.973 & 4.885 \\
KSA       & 0.686 & 0.581 & 0.496 & 0.420 & 0.912 & 0.913 & 0.912 & 4.936 & 4.860 & 4.957 & 4.970 & 4.792 \\
Lebanon   & 0.698 & 0.597 & 0.519 & 0.448 & 0.913 & 0.917 & 0.915 & 4.935 & 4.878 & 4.961 & 4.943 & 4.782 \\
Libya     & 0.720 & 0.628 & 0.554 & 0.486 & 0.922 & 0.924 & 0.923 & 4.870 & 4.893 & 4.991 & 4.912 & 4.758 \\
Morocco   & 0.735 & 0.641 & 0.562 & 0.490 & 0.923 & 0.925 & 0.924 & 4.911 & 4.895 & 4.980 & 4.915 & 4.790 \\
Palestine & 0.704 & 0.609 & 0.536 & 0.468 & 0.907 & 0.914 & 0.910 & 4.923 & 4.923 & 4.951 & 4.931 & 4.813 \\
Sudan     & 0.745 & 0.655 & 0.582 & 0.511 & 0.932 & 0.935 & 0.933 & 4.874 & 4.896 & 4.957 & 4.926 & 4.770 \\
Syria     & 0.628 & 0.514 & 0.428 & 0.353 & 0.890 & 0.886 & 0.888 & 4.896 & 4.857 & 4.956 & 4.936 & 4.757 \\
Tunisia   & 0.614 & 0.496 & 0.406 & 0.327 & 0.887 & 0.892 & 0.889 & 4.853 & 4.849 & 4.956 & 4.911 & 4.707 \\
UAE       & 0.722 & 0.632 & 0.557 & 0.486 & 0.922 & 0.922 & 0.922 & 4.929 & 4.875 & 4.933 & 4.926 & 4.753 \\
Yemen     & 0.657 & 0.552 & 0.469 & 0.394 & 0.895 & 0.898 & 0.896 & 4.829 & 4.878 & 4.935 & 4.858 & 4.707 \\

\midrule
\rowcolor{headergreen}
\multicolumn{13}{c}{\textbf{Best Arabic-Centric Model: ALLaM-7B}} \\

Algeria   & 0.623 & 0.507 & 0.420 & 0.343 & 0.892 & 0.884 & 0.888 & 4.091 & 4.482 & 4.745 & 4.465 & 3.988 \\
Egypt     & 0.585 & 0.467 & 0.386 & 0.316 & 0.870 & 0.871 & 0.871 & 4.356 & 4.427 & 4.515 & 4.598 & 4.080 \\
Jordan    & 0.740 & 0.657 & 0.589 & 0.525 & 0.914 & 0.914 & 0.914 & 4.513 & 4.636 & 4.740 & 4.755 & 4.291 \\
KSA       & 0.649 & 0.548 & 0.470 & 0.399 & 0.889 & 0.889 & 0.889 & 4.451 & 4.562 & 4.647 & 4.702 & 4.200 \\
Lebanon   & 0.678 & 0.579 & 0.506 & 0.438 & 0.900 & 0.898 & 0.899 & 4.179 & 4.524 & 4.707 & 4.555 & 4.070 \\
Libya     & 0.684 & 0.591 & 0.519 & 0.452 & 0.899 & 0.898 & 0.899 & 4.079 & 4.488 & 4.642 & 4.381 & 3.954 \\
Morocco   & 0.721 & 0.629 & 0.554 & 0.485 & 0.913 & 0.914 & 0.914 & 4.270 & 4.569 & 4.851 & 4.351 & 4.109 \\
Palestine & 0.661 & 0.564 & 0.491 & 0.422 & 0.887 & 0.893 & 0.890 & 4.455 & 4.561 & 4.683 & 4.724 & 4.203 \\
Sudan     & 0.699 & 0.609 & 0.540 & 0.476 & 0.903 & 0.907 & 0.905 & 4.296 & 4.596 & 4.691 & 4.426 & 4.161 \\
Syria     & 0.586 & 0.473 & 0.390 & 0.316 & 0.874 & 0.867 & 0.870 & 4.287 & 4.454 & 4.673 & 4.542 & 4.080 \\
Tunisia   & 0.570 & 0.446 & 0.358 & 0.282 & 0.866 & 0.867 & 0.866 & 3.844 & 4.302 & 4.502 & 4.058 & 3.724 \\
UAE       & 0.666 & 0.571 & 0.496 & 0.427 & 0.892 & 0.891 & 0.891 & 4.200 & 4.526 & 4.643 & 4.333 & 4.043 \\
Yemen     & 0.629 & 0.525 & 0.445 & 0.372 & 0.883 & 0.885 & 0.884 & 4.203 & 4.463 & 4.602 & 4.415 & 4.069 \\

\midrule
\rowcolor{headergold}
\multicolumn{13}{c}{\textbf{Best Multilingual Model (0-shot): Gemma-2-9B}} \\

Algeria   & 0.535 & 0.401 & 0.307 & 0.229 & 0.864 & 0.861 & 0.862 & 3.123 & 2.914 & 3.033 & 3.180 & 2.848 \\
Egypt     & 0.560 & 0.429 & 0.335 & 0.260 & 0.877 & 0.874 & 0.875 & 3.653 & 3.276 & 3.565 & 3.770 & 3.280 \\
Jordan    & 0.688 & 0.586 & 0.505 & 0.429 & 0.917 & 0.913 & 0.915 & 3.889 & 3.502 & 3.866 & 4.107 & 3.513 \\
KSA       & 0.627 & 0.509 & 0.421 & 0.344 & 0.890 & 0.888 & 0.889 & 3.706 & 3.477 & 3.706 & 3.877 & 3.379 \\
Lebanon   & 0.607 & 0.492 & 0.406 & 0.332 & 0.883 & 0.884 & 0.883 & 3.306 & 3.153 & 3.376 & 3.463 & 3.074 \\
Libya     & 0.635 & 0.531 & 0.453 & 0.382 & 0.887 & 0.888 & 0.887 & 3.335 & 3.186 & 3.251 & 3.572 & 3.056 \\
Morocco   & 0.531 & 0.419 & 0.336 & 0.266 & 0.832 & 0.857 & 0.843 & 2.782 & 2.710 & 3.008 & 2.891 & 2.609 \\
Palestine & 0.618 & 0.501 & 0.415 & 0.338 & 0.883 & 0.885 & 0.884 & 3.764 & 3.443 & 3.646 & 3.927 & 3.406 \\
Sudan     & 0.668 & 0.557 & 0.473 & 0.397 & 0.910 & 0.909 & 0.909 & 3.570 & 3.296 & 3.500 & 3.661 & 3.209 \\
Syria     & 0.548 & 0.422 & 0.333 & 0.258 & 0.868 & 0.861 & 0.864 & 3.434 & 3.171 & 3.502 & 3.614 & 3.139 \\
Tunisia   & 0.497 & 0.361 & 0.265 & 0.190 & 0.853 & 0.854 & 0.854 & 2.844 & 2.702 & 2.693 & 2.813 & 2.653 \\
UAE       & 0.651 & 0.545 & 0.459 & 0.382 & 0.903 & 0.898 & 0.900 & 3.471 & 3.396 & 3.608 & 3.553 & 3.200 \\
Yemen     & 0.606 & 0.493 & 0.407 & 0.330 & 0.886 & 0.886 & 0.886 & 3.553 & 3.362 & 3.573 & 3.589 & 3.240 \\

\midrule
\rowcolor{headerblue}
\multicolumn{13}{c}{\textbf{Best Multilingual Model (SFT): Gemma-2-9B}} \\

Algeria   & 0.497 & 0.397 & 0.320 & 0.253 & 0.851 & 0.820 & 0.835 & 2.123 & 3.131 & 3.812 & 3.074 & 2.615 \\
Egypt     & 0.486 & 0.386 & 0.312 & 0.248 & 0.850 & 0.824 & 0.837 & 2.117 & 3.067 & 3.778 & 3.297 & 2.665 \\
Jordan    & 0.596 & 0.524 & 0.463 & 0.405 & 0.878 & 0.848 & 0.863 & 2.157 & 3.180 & 3.939 & 3.498 & 2.808 \\
KSA       & 0.538 & 0.450 & 0.380 & 0.319 & 0.864 & 0.838 & 0.851 & 2.119 & 3.221 & 3.962 & 3.268 & 2.719 \\
Lebanon   & 0.549 & 0.465 & 0.399 & 0.342 & 0.862 & 0.835 & 0.848 & 2.087 & 3.183 & 3.887 & 3.031 & 2.638 \\
Libya     & 0.557 & 0.471 & 0.405 & 0.344 & 0.860 & 0.833 & 0.846 & 2.093 & 3.074 & 3.767 & 3.088 & 2.642 \\
Morocco   & 0.554 & 0.468 & 0.401 & 0.339 & 0.865 & 0.837 & 0.851 & 2.101 & 3.133 & 3.794 & 3.081 & 2.633 \\
Palestine & 0.552 & 0.469 & 0.404 & 0.344 & 0.866 & 0.841 & 0.853 & 2.081 & 3.077 & 3.890 & 3.281 & 2.679 \\
Sudan     & 0.590 & 0.514 & 0.451 & 0.392 & 0.872 & 0.845 & 0.858 & 2.152 & 3.048 & 3.783 & 3.200 & 2.674 \\
Syria     & 0.465 & 0.373 & 0.303 & 0.243 & 0.852 & 0.822 & 0.837 & 2.124 & 3.100 & 3.749 & 3.040 & 2.590 \\
Tunisia   & 0.467 & 0.363 & 0.285 & 0.221 & 0.842 & 0.818 & 0.830 & 2.084 & 3.196 & 3.742 & 2.831 & 2.600 \\
UAE       & 0.534 & 0.454 & 0.388 & 0.329 & 0.867 & 0.839 & 0.853 & 2.118 & 3.031 & 3.741 & 3.059 & 2.584 \\
Yemen     & 0.539 & 0.455 & 0.386 & 0.325 & 0.863 & 0.836 & 0.849 & 2.155 & 3.077 & 3.789 & 3.167 & 2.642 \\

\bottomrule
\end{tabular}

\caption{Country-wise Dialect-to-MSA translation performance under \emph{Context: Country + Region}, reporting results for the best-performing model in each category: proprietary (GPT-5), multilingual (Gemma-2-9B), Arabic-centric (ALLaM-7B), and multilingual SFT (Gemma-2-9B SFT). We report BLEU, BERTScore, and LLM-as-a-judge scores on a 1--5 scale.}
\label{tab:country_dialect_msa}
\end{table*}

\begin{table*}[!h]
\centering
\setlength{\tabcolsep}{3.5pt}
\begin{tabular}{l | cccc | ccc | ccccc}
\toprule
\multirow{2}{*}{\textbf{Region}} &
\multicolumn{4}{c|}{\textbf{BLEU}} &
\multicolumn{3}{c|}{\textbf{BERTScore}} &
\multicolumn{5}{c}{\textbf{LLM-as-a-judge (1--5)}} \\
\cline{2-13}\\
& B1 & B2 & B3 & B4 & P & R & F1 &
Adeq. & Flu. & Reg. & Term. & Overall \\
\midrule

\rowcolor{headerpink}
\multicolumn{13}{c}{\textbf{Best Proprietary Model: GPT-5}} \\

Gulf         & 0.607 & 0.481 & 0.386 & 0.304 & 0.884 & 0.886 & 0.885 & 4.917 & 4.576 & 4.232 & 4.914 & 4.318 \\
Levant       & 0.579 & 0.448 & 0.349 & 0.266 & 0.874 & 0.882 & 0.878 & 4.947 & 4.832 & 4.860 & 4.964 & 4.721 \\
Nile River   & 0.627 & 0.500 & 0.406 & 0.325 & 0.890 & 0.891 & 0.891 & 4.934 & 4.734 & 4.731 & 4.947 & 4.589 \\
North Africa & 0.562 & 0.425 & 0.324 & 0.241 & 0.866 & 0.872 & 0.869 & 4.869 & 4.655 & 4.720 & 4.882 & 4.467 \\

\midrule
\rowcolor{headergreen}
\multicolumn{13}{c}{\textbf{Best Arabic-Centric Model: ALLaM-7B-Instruct}} \\

Gulf         & 0.544 & 0.412 & 0.318 & 0.241 & 0.862 & 0.861 & 0.861 & 4.211 & 3.976 & 2.972 & 4.389 & 3.515 \\
Levant       & 0.515 & 0.379 & 0.283 & 0.205 & 0.858 & 0.858 & 0.858 & 4.144 & 3.883 & 3.629 & 4.324 & 3.655 \\
Nile River   & 0.539 & 0.400 & 0.308 & 0.231 & 0.859 & 0.858 & 0.859 & 4.075 & 3.870 & 3.320 & 4.137 & 3.418 \\
North Africa & 0.428 & 0.286 & 0.197 & 0.132 & 0.822 & 0.814 & 0.818 & 4.312 & 3.306 & 2.253 & 4.089 & 3.055 \\

\midrule
\rowcolor{headergold}
\multicolumn{13}{c}{\textbf{Best Multilingual Model (0-shot): Qwen-3-8B}} \\

Gulf         & 0.466 & 0.327 & 0.232 & 0.159 & 0.835 & 0.838 & 0.837 & 3.544 & 3.079 & 1.481 & 3.534 & 2.435 \\
Levant       & 0.389 & 0.247 & 0.159 & 0.100 & 0.818 & 0.817 & 0.818 & 3.526 & 2.798 & 1.593 & 3.513 & 2.445 \\
Nile River   & 0.423 & 0.277 & 0.186 & 0.121 & 0.826 & 0.822 & 0.824 & 3.284 & 2.633 & 1.576 & 3.279 & 2.309 \\
North Africa & 0.371 & 0.233 & 0.147 & 0.094 & 0.805 & 0.797 & 0.801 & 3.274 & 2.504 & 1.421 & 3.039 & 2.218 \\

\midrule
\rowcolor{headerblue}
\multicolumn{13}{c}{\textbf{Best Multilingual Model (SFT): Gemma-2-9B-it}} \\

Gulf         & 0.425 & 0.312 & 0.230 & 0.165 & 0.829 & 0.807 & 0.818 & 2.091 & 2.681 & 2.058 & 3.105 & 2.232 \\
Levant       & 0.379 & 0.261 & 0.180 & 0.121 & 0.819 & 0.797 & 0.808 & 2.077 & 2.570 & 2.443 & 3.057 & 2.297 \\
Nile River   & 0.422 & 0.310 & 0.229 & 0.166 & 0.826 & 0.803 & 0.814 & 2.066 & 2.544 & 2.414 & 3.006 & 2.258 \\
North Africa & 0.351 & 0.240 & 0.162 & 0.109 & 0.804 & 0.779 & 0.791 & 2.052 & 2.312 & 1.896 & 2.653 & 2.077 \\

\bottomrule
\end{tabular}
\caption{Region-wise MSA-to-Dialect translation performance under \emph{Context: Country + Region}, reporting results for the best-performing model in each category: proprietary (GPT-5), multilingual (Qwen-3-8B), Arabic-centric (ALLaM-7B), and multilingual SFT (Gemma-2-9B). We report BLEU, BERTScore, and LLM-as-a-judge scores on a 1--5 scale.}
\label{tab:region_msa_dialect}
\end{table*}

\begin{table*}[!h]
\centering
\setlength{\tabcolsep}{3.pt}
\begin{tabular}{l | cccc | ccc | ccccc}
\toprule
\multirow{2}{*}{\textbf{Region}} &
\multicolumn{4}{c|}{\textbf{BLEU}} &
\multicolumn{3}{c|}{\textbf{BERTScore}} &
\multicolumn{5}{c}{\textbf{LLM-as-a-judge (1--5)}} \\
\cline{2-13}\\
& B1 & B2 & B3 & B4 & P & R & F1 &
Adeq. & Flu. & Reg. & Term. & Overall \\
\midrule

\rowcolor{headerpink}
\multicolumn{13}{c}{\textbf{Best Proprietary Model: GPT-5}} \\

Gulf         & 0.689 & 0.589 & 0.508 & 0.434 & 0.910 & 0.911 & 0.910 & 4.898 & 4.871 & 4.942 & 4.917 & 4.750 \\
Levant       & 0.700 & 0.602 & 0.525 & 0.454 & 0.911 & 0.913 & 0.912 & 4.935 & 4.897 & 4.957 & 4.946 & 4.811 \\
Nile River   & 0.684 & 0.581 & 0.500 & 0.427 & 0.914 & 0.916 & 0.915 & 4.898 & 4.889 & 4.953 & 4.925 & 4.776 \\
North Africa & 0.678 & 0.573 & 0.490 & 0.416 & 0.908 & 0.910 & 0.909 & 4.883 & 4.876 & 4.971 & 4.914 & 4.750 \\

\midrule
\rowcolor{headergreen}
\multicolumn{13}{c}{\textbf{Best Arabic-Centric Model: ALLaM-7B-Instruct}} \\

Gulf         & 0.648 & 0.548 & 0.471 & 0.400 & 0.888 & 0.889 & 0.888 & 4.281 & 4.516 & 4.630 & 4.478 & 4.102 \\
Levant       & 0.666 & 0.569 & 0.495 & 0.426 & 0.894 & 0.893 & 0.893 & 4.364 & 4.545 & 4.701 & 4.646 & 4.164 \\
Nile River   & 0.641 & 0.537 & 0.462 & 0.394 & 0.887 & 0.889 & 0.888 & 4.326 & 4.510 & 4.601 & 4.514 & 4.119 \\
North Africa & 0.650 & 0.544 & 0.464 & 0.391 & 0.893 & 0.891 & 0.892 & 4.076 & 4.463 & 4.691 & 4.317 & 3.948 \\

\midrule
\rowcolor{headergold}
\multicolumn{13}{c}{\textbf{Best Multilingual Model (0-shot): Gemma-2-9B-it}} \\

Gulf         & 0.628 & 0.516 & 0.429 & 0.353 & 0.893 & 0.891 & 0.892 & 3.573 & 3.410 & 3.628 & 3.669 & 3.270 \\
Levant       & 0.616 & 0.501 & 0.416 & 0.340 & 0.888 & 0.886 & 0.887 & 3.607 & 3.322 & 3.605 & 3.787 & 3.290 \\
Nile River   & 0.613 & 0.492 & 0.403 & 0.327 & 0.893 & 0.891 & 0.892 & 3.612 & 3.286 & 3.533 & 3.716 & 3.245 \\
North Africa & 0.548 & 0.426 & 0.338 & 0.265 & 0.858 & 0.864 & 0.861 & 3.014 & 2.871 & 2.995 & 3.105 & 2.785 \\

\midrule
\rowcolor{headerblue}
\multicolumn{13}{c}{\textbf{Best Multilingual Model (SFT): Gemma-2-9B-it}} \\

Gulf         & 0.537 & 0.453 & 0.385 & 0.324 & 0.865 & 0.838 & 0.851 & 2.130 & 3.107 & 3.827 & 3.162 & 2.647 \\
Levant       & 0.541 & 0.458 & 0.393 & 0.334 & 0.865 & 0.837 & 0.850 & 2.114 & 3.135 & 3.866 & 3.219 & 2.681 \\
Nile River   & 0.537 & 0.449 & 0.380 & 0.319 & 0.861 & 0.834 & 0.847 & 2.134 & 3.058 & 3.780 & 3.250 & 2.670 \\
North Africa & 0.519 & 0.425 & 0.353 & 0.289 & 0.855 & 0.827 & 0.840 & 2.101 & 3.134 & 3.780 & 3.020 & 2.622 \\

\bottomrule
\end{tabular}
\caption{Region-wise Dialect-to-MSA translation performance under \emph{Context: Country + Region}, reporting results for the best-performing model in each category: proprietary (GPT-5), multilingual (Gemma-2-9B), Arabic-centric (ALLaM-7B), and multilingual SFT (Gemma-2-9B SFT). We report BLEU, BERTScore, and LLM-as-a-judge scores on a 1--5 scale.}
\label{tab:region_dialect_msa}
\end{table*}

\section{Agreement Between Human Evaluation and LLM-as-a-judge}
\label{sec:human_eval-agreement}

\begin{figure*}[!h]
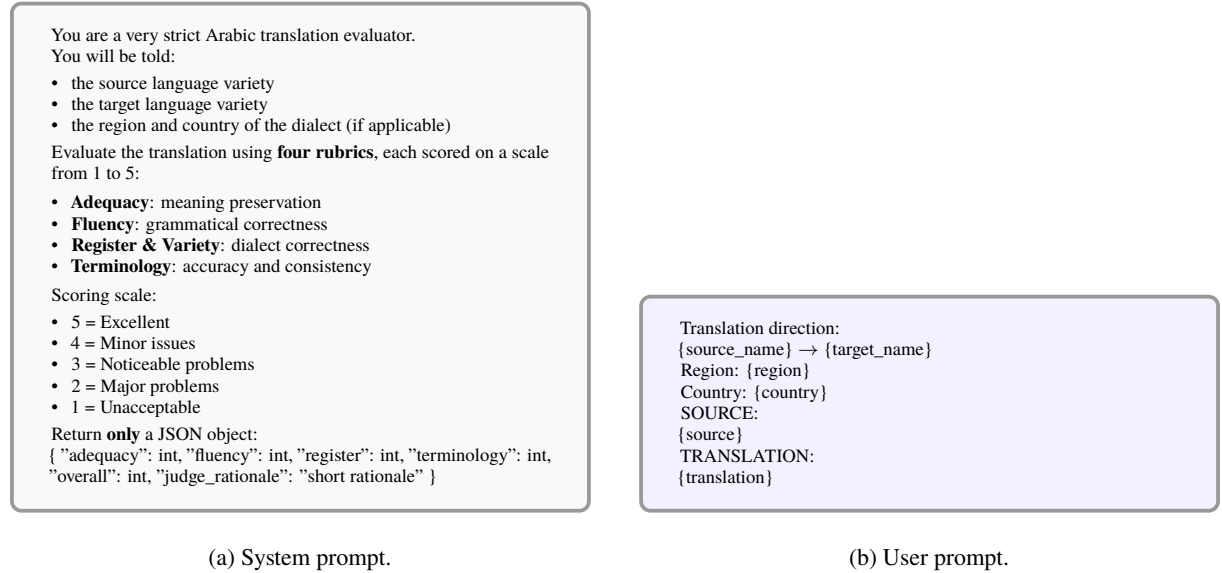

  \centering

  \begin{subfigure}[t]{0.48\textwidth}
    \centering
    \begin{tcolorbox}[colback=gray!5, colframe=black!40]
\scriptsize
You are a very strict Arabic translation evaluator.

You will be told:
\begin{itemize}[noitemsep, topsep=2pt, leftmargin=*]
    \item the source language variety
    \item the target language variety
    \item the region and country of the dialect (if applicable)
\end{itemize}

Evaluate the translation using \textbf{four rubrics}, each scored on a scale from 1 to 5:

\begin{itemize}[noitemsep, topsep=2pt, leftmargin=*]
    \item \textbf{Adequacy}: meaning preservation
    \item \textbf{Fluency}: grammatical correctness
    \item \textbf{Register \& Variety}: dialect correctness
    \item \textbf{Terminology}: accuracy and consistency
\end{itemize}

Scoring scale:
\begin{itemize}[noitemsep, topsep=2pt, leftmargin=*]
    \item 5 = Excellent
    \item 4 = Minor issues
    \item 3 = Noticeable problems
    \item 2 = Major problems
    \item 1 = Unacceptable
\end{itemize}

Return \textbf{only} a JSON object:

\{
  "adequacy": int,
  "fluency": int,
  "register": int,
  "terminology": int,
  "overall": int,
  "judge\_rationale": "short rationale"
\}
    \end{tcolorbox}
    \caption{System prompt.}
  \end{subfigure}
  \hfill
  \begin{subfigure}[t]{0.48\textwidth}
    \centering
    \begin{tcolorbox}[colback=blue!5, colframe=black!40]
\scriptsize
Translation direction: \\
\{source\_name\} $\rightarrow$ \{target\_name\}

Region: \{region\} \\
Country: \{country\}

SOURCE: \\
\{source\}

TRANSLATION: \\
\{translation\}
    \end{tcolorbox}
    \caption{User prompt.}
  \end{subfigure}

  \caption{Prompting strategy for LLM-based evaluation: a fixed system prompt defines criteria and scoring, while the user prompt provides translation direction, regional metadata, and the source--translation pair.}
  \label{fig:appendix-prompt-judge}
\end{figure*}

\autoref{fig:appendix-prompt-judge} shows the prompt used for LLM-based evaluation, where a language model is instructed to act as a strict Arabic translation judge and score outputs along multiple linguistic and cultural dimensions.
To assess the reliability of the LLM-as-a-judge evaluation, we analyze its agreement with human judgments on a subset of the Moroccan, Egyptian, Lebanese, and Emirati samples.

\autoref{tab:human_llm_agreement} reports the agreement between human judgments and LLM-as-a-Judge scores for the adequacy and fluency rubric. We compute Mean Absolute Difference (MAD) to quantify the average absolute deviation between human and LLM-as-a-Judge scores, and Accuracy@1 to measure the proportion of instances in which the LLM score falls within one point of the averaged human score. Overall, MAD values remain mostly below 1 across the four considered dialects. 

Notably, the difference between the manually assigned ratings and the LLM-as-judge ratings varies for the outputs of the three considered models. More specifically, the MAD for GPT-5 and Allam-7B-Instruct are low, while the MADs are a bit higher for Qwen3-8B which generates worse translations as per \autoref{tab:msa_dialect_main}. This hints that the LLM-as-judge ratings are generally useful, yet, further investigations are needed to assess their reliability as robust quality estimators.

\begin{table*}[!h]
\centering
\small
\begin{tabular}{l|l|cc|cc}
\hline
& & \multicolumn{2}{c|}{\textbf{\textit{Adequacy}}} & \multicolumn{2}{c}{\textbf{\textit{Fluency}}} \\
\hline
\textbf{Direction} & \textbf{Model} & \textbf{MAD $\downarrow$ [0-4]} & \textbf{Acc@1 $\uparrow$ [0-1]} & \textbf{MAD $\downarrow$ [0-4]} & \textbf{Acc@1 $\uparrow$ [0-1]} \\
\hline

\multicolumn{6}{c}{\textbf{Moroccan}} \\
\hline
MSA $\rightarrow$ Dial & GPT-5          & 0.04 & 1.00 & 0.40 & 1.00 \\
MSA $\rightarrow$ Dial & Allam-7B-Inst  & 0.48 & 0.96 & 0.36 & 1.00 \\
MSA $\rightarrow$ Dial & Qwen3-8B       & 1.52 & 0.52 & 0.48 & 1.00 \\
\hline
Dial $\rightarrow$ MSA & GPT-5          & 0.00 & 1.00 & 0.00 & 1.00 \\
Dial $\rightarrow$ MSA & Allam-7B-Inst  & 0.84 & 0.96 & 0.64 & 0.80 \\
Dial $\rightarrow$ MSA & Qwen3-8B       & 1.28 & 0.68 & 1.96 & 0.24 \\
\hline

\multicolumn{6}{c}{\textbf{Egyptian}} \\
\hline
MSA $\rightarrow$ Dial & GPT-5          & 0.40 & 1.00 & 0.52 & 0.96 \\
MSA $\rightarrow$ Dial & Allam-7B-Inst  & 0.84 & 0.84 & 1.28 & 0.60 \\
MSA $\rightarrow$ Dial & Qwen3-8B       & 1.28 & 0.60 & 0.72 & 0.84 \\
\hline
Dial $\rightarrow$ MSA & GPT-5          & 0.08 & 1.00 & 0.64 & 0.80 \\
Dial $\rightarrow$ MSA & Allam-7B-Inst  & 0.44 & 0.92 & 0.76 & 0.84 \\
Dial $\rightarrow$ MSA & Qwen3-8B       & 1.04 & 0.68 & 0.92 & 0.76 \\
\hline

\multicolumn{6}{c}{\textbf{Lebanese}} \\
\hline
MSA $\rightarrow$ Dial & GPT-5          & 0.04 & 1.00 & 0.04 & 1.00 \\
MSA $\rightarrow$ Dial & Allam-7B-Inst  & 0.60 & 1.00 & 0.88 & 0.92 \\
MSA $\rightarrow$ Dial & Qwen3-8B       & 1.76 & 0.36 & 0.72 & 0.96 \\
\hline
Dial $\rightarrow$ MSA & GPT-5          & 0.04 & 1.00 & 0.04 & 1.00 \\
Dial $\rightarrow$ MSA & Allam-7B-Inst  & 0.40 & 1.00 & 0.20 & 1.00 \\
Dial $\rightarrow$ MSA & Qwen3-8B       & 1.16 & 0.60 & 1.48 & 0.44 \\
\hline

\multicolumn{6}{c}{\textbf{Emirati}} \\
\hline
MSA $\rightarrow$ Dial & GPT-5          & 0.04 & 1.00 & 0.76 & 0.84 \\
MSA $\rightarrow$ Dial & Allam-7B-Inst  & 0.60 & 0.80 & 1.68 & 0.60 \\
MSA $\rightarrow$ Dial & Qwen3-8B       & 0.92 & 0.80 & 3.40 & 0.20 \\
\hline
Dial $\rightarrow$ MSA & GPT-5          & 0.32 & 0.96 & 0.20 & 0.96 \\
Dial $\rightarrow$ MSA & Allam-7B-Inst  & 0.40 & 1.00 & 0.56 & 1.00 \\
Dial $\rightarrow$ MSA & Qwen3-8B       & 1.80 & 0.44 & 1.00 & 0.80 \\
\hline

\end{tabular}
\caption{Agreement between human judgments and LLM-as-a-judge scores for different models on the Moroccan, Egyptian, Lebanese, and Emirati datasets sample. Mean Absolute Difference (MAD) ranges from 0 to 4, with lower values indicating closer alignment between human and LLM-as-a-judge scores. Accuracy@1  measures the proportion of instances in which the LLM score falls within one point of the manually-assigned score.}
\label{tab:human_llm_agreement}
\end{table*}

\section{Task 3 - Dialect Steering Analysis}
\label{sec:dialect_steering_appendix}

{\setlength{\tabcolsep}{0.35em}
\renewcommand{\arraystretch}{1.26}
\begin{table*}[!h]
\centering
\small
\begin{tabular}{l|cc|cc|cc}
\toprule
\textbf{Country/Region} &
\multicolumn{2}{c|}{\cellcolor{headerpink}\shortstack{\textbf{Best Proprietary}\\\textbf{(GPT-5)}}} &
\multicolumn{2}{c|}{\cellcolor{headergold}\shortstack{\textbf{Best Multilingual}\\\textbf{(Gemma-2-9B-it)}}} &
\multicolumn{2}{c}{\cellcolor{headergreen}\shortstack{\textbf{Best Arabic-centric}\\\textbf{(ALLaM-7B)}}} \\
 & \textbf{Judge} & \textbf{GlotLID} & \textbf{Judge} & \textbf{GlotLID} & \textbf{Judge} & \textbf{GlotLID} \\
\midrule
Algeria  & 0.963 & 0.045 & 0.568 & 0.082 & 0.819 & 0.049 \\
Egypt    & 0.962 & 0.908 & 0.695 & 0.682 & 0.865 & 0.816 \\
Jordan   & 0.969 & 0.552 & 0.656 & 0.172 & 0.889 & 0.433 \\
KSA      & 0.959 & 0.511 & 0.673 & 0.217 & 0.857 & 0.579 \\
Lebanon  & 0.959 & 0.729 & 0.629 & 0.279 & 0.845 & 0.664 \\
Libya    & 0.951 & 0.000 & 0.568 & 0.000 & 0.757 & 0.000 \\
Morocco  & 0.968 & 0.988 & 0.551 & 0.182 & 0.799 & 0.660 \\
Palestine& 0.966 & 0.610 & 0.616 & 0.211 & 0.849 & 0.411 \\
Sudan    & 0.947 & 0.000 & 0.570 & 0.000 & 0.799 & 0.000 \\
Syria    & 0.952 & 0.733 & 0.625 & 0.224 & 0.815 & 0.545 \\
Tunisia  & 0.958 & 0.893 & 0.522 & 0.084 & 0.762 & 0.604 \\
UAE      & 0.947 & 0.000 & 0.592 & 0.000 & 0.803 & 0.000 \\
Yemen    & 0.917 & 0.000 & 0.586 & 0.000 & 0.744 & 0.000 \\
\midrule
\textbf{Gulf}         & 0.941 & 0.609 & 0.616 & 0.273 & 0.801 & 0.533 \\
\textbf{Levant}       & 0.963 & 0.646 & 0.633 & 0.220 & 0.853 & 0.506 \\
\textbf{Nile River}   & 0.955 & 0.731 & 0.634 & 0.595 & 0.833 & 0.695 \\
\textbf{North Africa} & 0.960 & 0.929 & 0.553 & 0.284 & 0.786 & 0.693 \\
\bottomrule
\end{tabular}
\caption{Dialect steering by country (zero-shot): best model per family. GlotLID is strict ISO-code accuracy; region rows use the macro-region mapping from \citet{bhatti2025beyond}.}
\label{tab:dialect-steering-by-country}
\end{table*}
}

{\setlength{\tabcolsep}{0.35em}
\renewcommand{\arraystretch}{0.7}
\begin{table*}[!h]
\centering
\footnotesize
\resizebox{\textwidth}{!}{%
\begin{tabular}{l p{0.20\textwidth} p{0.32\textwidth} p{0.40\textwidth}}
\toprule
\textbf{Country} & \textbf{Sample ID} & \textbf{Prompt excerpt} & \textbf{Model continuations (zero-shot) + predicted GlotLID} \\
\midrule
UAE & \texttt{uae-\allowbreak 15-\allowbreak traditions-\allowbreak during-\allowbreak religous-\allowbreak holidays-\allowbreak dialect} & \begin{arab}هزيم: كل عام وأنت بخير يا سالم!\end{arab}\par \begin{arab}سالم: وأنت بخير يا هزيم، ما أجمل أجواء العيد بين الأهل والأصدقاء.\end{arab}\par \begin{arab}هزيم: في كل عيد أحرص ابارك بنفس العبارة التي تعلمتها من والدي.\end{arab}\par \begin{arab}سالم: نعم، هذه العبارة تملأ القلب بالفرح وتعتبر من عاداتنا.\end{arab}\par \begin{arab}هزيم: من الجميل أن نحافظ على تقاليدنا وننقلها للأجيال القادمة.\end{arab}\par \begin{arab}سالم: ...\end{arab} & \textbf{GPT-5} (\texttt{ars}):\par \begin{arab}صح لسانك يا هزيم، ودامت أفراحكم، شو رايك عقب صلاة العيد نمر على بيوت الربع نسلّم ونعايدهم، وبالليل تتفضلون عندنا للقهوة والحلو والعيديات؟\end{arab}\par \textbf{ALLaM-7B} (\texttt{ars}):\par \begin{arab}والله يا سالم، العيد ما يحلا إلا بوجودك، وكل عام وأنت بخير وعيدك مبارك يا الغالي.\end{arab}\par \textbf{Gemma-2-9B-it} (\texttt{ars}):\par \begin{arab}كل عام وانت بالف خير يا هزيم، وربي يبارك فيك ويحفظك.\end{arab} \\
Morocco & \texttt{morocco-\allowbreak 2-\allowbreak traditions-\allowbreak before-\allowbreak religous-\allowbreak holidays-\allowbreak dialect} & \begin{arab}فاطمة: رمضان على الأبواب، هل بدأتِ الاستعدادات يا ليلى؟\end{arab}\par \begin{arab}ليلى: نعم، في هذا الوقت تبدأ الأمهات بتحضير ما يلزم للشهر الكريم.\end{arab}\par \begin{arab}فاطمة: رائحة الشباكية والسلو تنتشر في كل بيت تقريبًا.\end{arab}\par \begin{arab}ليلى: صحيح، كل أسرة تحرص على تجهيز كميات تكفي لأيام الشهر الفضيل.\end{arab}\par \begin{arab}فاطمة: الجو كله يصبح مليئًا بالفرح والنشاط في المطبخ.\end{arab}\par \begin{arab}ليلى: ...\end{arab} & \textbf{GPT-5} (\texttt{ary}):\par \begin{arab}إي بصح، كنفيقو بكري نديرو الشباكية والسلو، ونقسمو الوقت بين الطياب والعبادة، وما نساوش الحريرة والتمر والشاي، وكنشريو المقاضي قبل الزحام باش نتهناو فالأيام اللّولة.\end{arab}\par \textbf{ALLaM-7B} (\texttt{ary}):\par \begin{arab}والله حتى كاينة أجواء زوينة في الشهر الكريم، الكل مجمعين على مائدة الإفطار وراجنين على تحضير الأكل.\end{arab}\par \textbf{Gemma-2-9B-it} (\texttt{apc}):\par \begin{arab}والليلة هادفنا نعمل السُّمن\end{arab} \\
Syria & \texttt{syria-\allowbreak 10-\allowbreak gift-\allowbreak giving-\allowbreak practices-\allowbreak the-\allowbreak occasion-\allowbreak dialect} & \begin{arab}ليلى: سمعت أن جمال قدّم هدية كبيرة لأخيه مؤخراً.\end{arab}\par \begin{arab}سامر: نعم، لقد اختار قطعة أثاث مميزة جداً.\end{arab}\par \begin{arab}ليلى: هل كان هناك مناسبة خاصة لذلك؟\end{arab}\par \begin{arab}سامر: بالتأكيد، في مجتمعنا هناك تقاليد مرتبطة بمثل هذه الهدايا.\end{arab}\par \begin{arab}ليلى: الهدايا عادةً ما تعكس المناسبة وأهميتها عندنا.\end{arab}\par \begin{arab}سامر: ...\end{arab} & \textbf{GPT-5} (\texttt{apc}):\par \begin{arab}صح، هيك الهدايا بتكون عادة بمناسبات متل الخطبة أو فتح بيت جديد، وعيب الواحد يروح فاضي الإيدين، فبيعطي شي إلو قيمة وبيضل ذكرى؛ وجمال اختار خزانة أنتيك بتجنّن لأن أخوه هالفترة عم يجهّز بيته.\end{arab}\par \textbf{ALLaM-7B} (\texttt{apc}):\par \begin{arab}هيك شي، بهدايا كبيرة متل هيك بيضلوا فترة طويلة عند الناس وبيحكوا فيها، خاصة إذا كانت قطعة أثاث مميزة مثل ما قال سامر.\end{arab}\par \textbf{Gemma-2-9B-it} (\texttt{arb}):\par \begin{arab}بالتأكيد، هدايا جمال كانت تتناسب مع أهمية المناسبة.\end{arab} \\
\bottomrule
\end{tabular}%
}
\caption{Case studies for dialect steering (zero-shot), focusing on UAE (Gulf), Morocco (Darija), and Syria (Levant). All Arabic excerpts are rendered with babel (Arabic); GlotLID codes are the strict-code predictions for each continuation.}
\label{tab:dialect-steering-cases}
\end{table*}
}

As mentioned in Section~\ref{sec:dialect-steering}, models show variable ability to use the intended dialect in dialogue. This is partly due to misalignment between GlotLID labels and our country-level dialects. For instance, a strict ISO-code GlotLID accuracy of 0 is realized for Gulf prompts, as shown in \autoref{tab:dialect-steering-by-country}, even when the continuation is clearly colloquial and regionally plausible.\\\\\\

{\setlength{\tabcolsep}{0.35em}
\renewcommand{\arraystretch}{1}
\begin{table*}[!h]
\centering
\begin{tabular}{l|cc|cc|cc}
\toprule
\textbf{Country/Region} &
\multicolumn{2}{c|}{\cellcolor{headerblue}\shortstack{\textbf{SFT}\\\textbf{(Llama-3.1-8B-Instruct)}}} &
\multicolumn{2}{c|}{\cellcolor{headerblue}\shortstack{\textbf{SFT}\\\textbf{(Qwen-3-8B)}}} &
\multicolumn{2}{c}{\cellcolor{headerblue}\shortstack{\textbf{SFT}\\\textbf{(Gemma-2-9B)}}} \\
 & \textbf{Judge} & \textbf{GlotLID} & \textbf{Judge} & \textbf{GlotLID} & \textbf{Judge} & \textbf{GlotLID} \\
\midrule
Algeria    & 0.466 & 0.016 & 0.529 & 0.008 & 0.560 & 0.012 \\
Egypt      & 0.514 & 0.435 & 0.562 & 0.238 & 0.616 & 0.506 \\
Jordan     & 0.566 & 0.115 & 0.557 & 0.054 & 0.674 & 0.153 \\
KSA        & 0.556 & 0.145 & 0.582 & 0.213 & 0.645 & 0.187 \\
Lebanon    & 0.512 & 0.210 & 0.533 & 0.048 & 0.584 & 0.175 \\
Libya      & 0.499 & 0.000 & 0.549 & 0.000 & 0.605 & 0.000 \\
Morocco    & 0.496 & 0.348 & 0.484 & 0.134 & 0.568 & 0.563 \\
Palestine  & 0.545 & 0.126 & 0.537 & 0.077 & 0.632 & 0.199 \\
Sudan      & 0.516 & 0.000 & 0.534 & 0.000 & 0.603 & 0.000 \\
Syria      & 0.498 & 0.170 & 0.530 & 0.103 & 0.604 & 0.194 \\
Tunisia    & 0.452 & 0.084 & 0.498 & 0.129 & 0.535 & 0.209 \\
UAE        & 0.528 & 0.000 & 0.559 & 0.000 & 0.620 & 0.000 \\
Yemen      & 0.525 & 0.000 & 0.515 & 0.000 & 0.589 & 0.000 \\
\midrule
\textbf{Gulf} & 0.536 & 0.159 & 0.551 & 0.197 & 0.618 & 0.200 \\
\textbf{Levant} & 0.534 & 0.152 & 0.540 & 0.068 & 0.627 & 0.179 \\
\textbf{Nile River} & 0.515 & 0.407 & 0.548 & 0.222 & 0.610 & 0.478 \\
\textbf{North Africa} & 0.478 & 0.361 & 0.514 & 0.233 & 0.566 & 0.574 \\
\bottomrule
\end{tabular}
\caption{Dialect steering by country: SFT models. GlotLID is strict ISO-code accuracy; region rows use the macro-region mapping from \citet{bhatti2025beyond}.}
\label{tab:dialect-steering-by-country-sft}
\end{table*}
}

This is most visible for the UAE split, where responses are predicted in the broad Gulf Arabic code (\texttt{afb}) rather than a UAE-exclusive dialect label, and for Saudi Arabia, where the same country label spans multiple major varieties (e.g., Najdi, Hijazi, and Gulf-adjacent Eastern speech).
\pagebreak
\newline

In both cases, models often predict a neighboring code (commonly Najdi \texttt{ars}), which strict exact-match scoring penalizes. We nevertheless report strict-code GlotLID because many downstream pipelines treat dialect as a discrete label, but we interpret it jointly with judged quality and the macro-region rows.
\pagebreak

We also note that GlotLID is imperfect and that dialect identification is not a single-label classification task \cite{keleg-magdy-2023-arabic}.

\paragraph{Case studies across Gulf, Darija, and Levant:} \autoref{tab:dialect-steering-cases} grounds the aggregate trends in concrete generations for UAE (Gulf), Morocco (Darija), and Syria (Levant). Two consistent phenomena stand out. First, code stability is hardest in the Gulf: even when models produce unmistakably colloquial Gulf continuations, GlotLID often assigns a neighboring label (frequently \texttt{ars}, Najdi) rather than the Gulf ISO label (\texttt{afb}) used for the UAE split, which explains the persistent strict-code zeros for UAE in \autoref{tab:dialect-steering-by-country}. This is not surprising given that Saudi Arabic is not a single uniform target in practice---Najdi and Hijazi are both prominent, and Eastern (Gulf-adjacent) speech shares many cues with UAE-style Gulf---so the UAE/KSA boundary is an especially fragile place to demand exact-code agreement. Second, the ``dialect'' target is not a single knob: Morocco behaves like a distinctive lexical style that supervision can amplify, whereas Syria is easier to keep fluent but more prone to drift toward pan-Levantine or even MSA-like realizations, especially when the content is generic.

Dialect control depends on label granularity and linguistic distinctiveness. A useful way to read the supervised results, therefore, is not as ``SFT always improves dialect'' but as ``SFT improves controllability and fluency, and it improves dialect identity when the target dialect has separable cues that the training signal reinforces.'' This aligns with the observed variation across regions, where some dialects benefit more from supervision than others. It also suggests that improvements in dialect steering depend on how clearly the target dialect can be distinguished from related varieties.